\documentclass[10pt,journal,compsoc]{IEEEtran}
\newif\ifpeerreview

\peerreviewfalse

\usepackage[nocompress]{cite}
\usepackage{url}
\usepackage{amsmath,amssymb,graphicx}
\usepackage{lipsum} 
\usepackage[switch]{lineno}
\usepackage[dvipsnames]{xcolor}

\usepackage[pagebackref,breaklinks]{hyperref}
\usepackage{multirow}
\usepackage{xcolor}
\usepackage{bm}
\usepackage{fnpos} 
\usepackage{overpic}

\usepackage{xspace}

\makeatletter
\DeclareRobustCommand\onedot{\futurelet\@let@token\@onedot}
\def\@onedot{\ifx\@let@token.\else.\null\fi\xspace}

\def\eg{\emph{e.g}\onedot} 
\def\ie{\emph{i.e}\onedot} 
 
\def\etc{\emph{etc}\onedot} 
 
\def\etal{\emph{et al}\onedot}
\makeatother

\newcommand{\paperID}{0006}

\title{Textureless Deformable Object Tracking \\with Invisible Markers}

 \author{Xianyuan~Li,~Yu~Guo,~Yubei~Tu,~Yu~Ji,~Yanchen~Liu,~Jinwei~Ye~\IEEEmembership{Senior~Member,~IEEE},~and~Changxi~Zheng
\IEEEcompsocitemizethanks{\IEEEcompsocthanksitem X. Li is with Tencent USA, New York,
 NY, 10018.
 \IEEEcompsocthanksitem Y. Guo is with George Mason University, Fairfax,
 VA, 22033.
 \IEEEcompsocthanksitem Y. Tu is with George Mason University, Fairfax,
 VA, 22033.
  \IEEEcompsocthanksitem Y. Ji is with LightThought LLC, New York,
 NY, 10018.
   \IEEEcompsocthanksitem Y. Liu is with Columbia University, New York,
 NY, 10027.
  \IEEEcompsocthanksitem J. Ye is with George Mason University, Fairfax,
 VA, 22033.
 \protect\\
E-mail: jinweiye@gmu.edu
 \IEEEcompsocthanksitem C. Zheng is with Columbia University, New York,
 NY, 10027. \protect\\
 E-mail: cxz@cs.columbia.edu}
}

\begin{document}

\IEEEtitleabstractindextext{%
\begin{abstract}
Tracking and reconstructing deformable objects with little texture is challenging due to the lack of features. Here we introduce ``invisible markers" for accurate and robust correspondence matching and tracking. Our markers are visible only under ultraviolet (UV) light. We build a novel imaging system for capturing videos of deformed objects under their original untouched appearance (which has little texture) and, simultaneously, with our markers. We develop an algorithm that first establishes accurate correspondences using video frames with markers, and then transfers them to the untouched textureless views as ground-truth labels. In this way, we are able to generate high-quality labeled data for training learning-based algorithms. We contribute a large real-world dataset, DOT, for tracking deformable objects with little or no texture. Our dataset has about one million video frames of various types of deformable objects. We provide ground truth tracked correspondences in both 2D and 3D. We benchmark state-of-the-art optical flow and deformable object tracking methods on our dataset. DOT poses challenges to these methods. By training networks on DOT, their performance on tracking deformable surface with little texture significantly improves, not only on our dataset,but also on other unseen data.

\end{abstract}

\begin{IEEEkeywords} 
Deformable Object Tracking, Deformable Surface Reconstruction, Invisible Light Imaging
\end{IEEEkeywords}
}

\maketitle

%
%
%

\IEEEraisesectionheading{
  \section{Introduction}\label{sec:introduction}
}

\IEEEPARstart{N}{umerous} applications, ranging from animation synthesis to robotic manipulation, need to track correspondences on deformable objects in order to understand their motion and deformation. Typical examples include cloth acquisition \cite{Hasler06VMV,White2007SIG} and gesture recognition \cite{zhou2019monocular,WANG2020SIG}.

Many algorithms rely on surface texture to establish correspondences~\cite{7328075,4668346,Ngo_2015_ICCV,Qi_2015_BMVC,Pizarro_2010_IJCV,6619043,7010934,7110605,Wang_2019_ICCV}. However, when an object has little or no texture---such as human skin or a piece of white paper---these methods all fall short. Some recent works train neural networks~\cite{8491013,cgf13125,Tsoli_2019_ICCV} to predict correspondences on textureless surfaces. Yet these methods are usually specific to a certain dataset. Cross-category generalization remains challenging. Furthermore, learning-based methods all face a chicken-and-egg problem: in order to obtain a dataset for training, one needs ground truth correspondences on textureless objects in the first place. A straightforward solution is to attach markers to explicitly introduce features. However, most markers would change the object's original appearance, making it hard to pair correspondences with the untouched textureless look.

In this paper, we introduce a novel type of ``invisible markers" that add features to a surface without changing its appearance under normal lighting conditions. Our makers are made with \emph{fluorescent dyes}, which are only visible under ultraviolet (UV) light, whereas invisible under normal light (in the visible spectrum). To capture the object in its original appearance, as well as with markers, we build a multi-view imaging system with UV lights: under UV lights, the markers would appear, providing rich features for accurate correspondence matching and tracking; under normal lights, the markers become invisible, allowing us to record the object's original appearance. The two types of lights are triggered in an interleaving fashion with a delay of a few milliseconds. In this way, videos with and without markers are synced.

Since the two types of videos are captured from different viewpoints, we transfer the tracked correspondences from videos with marker to the ones without. Correspondences are first matched using the marker videos for 3D reconstruction (among multiple views) and tracking (across time). We then devise a template-based method for registering the tracked correspondences onto the recovered 3D models. Finally, the correspondences are projected back into marker-free videos as ground truth labels. 

\begin{figure}[t]
\centering
\includegraphics[width=1.0\linewidth]{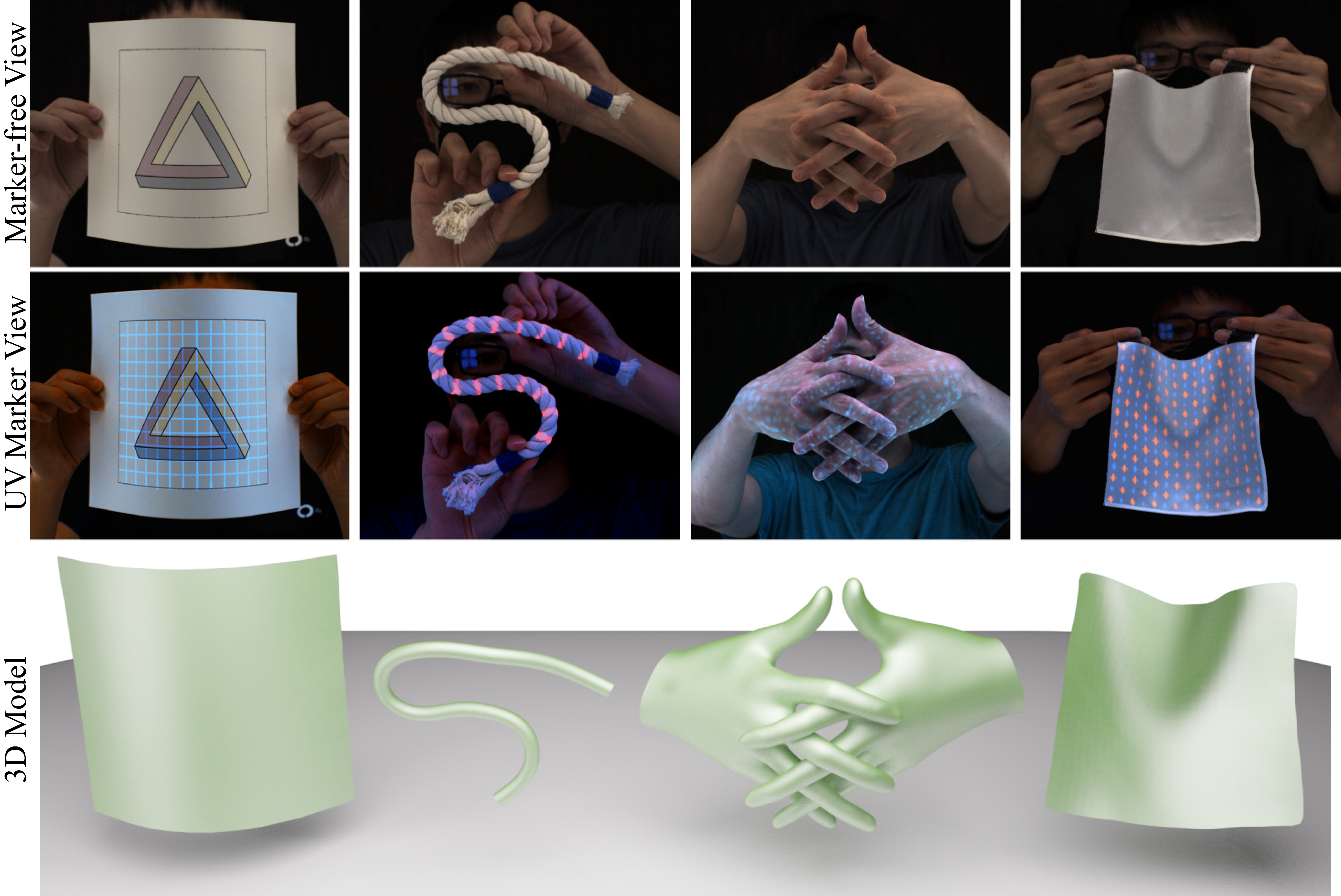}
\caption{Sample data of our deformable object tracking dataset (DOT). 
DOT has about one million video frames, featuring the deformation of various kinds of objects. For each deformation,
we provide multi-view video sequences with and without markers, recovered 3D models, and ground truth surface correspondences in both 2D and 3D.}
\label{fig:sample_data}
\end{figure}

Using our imaging system and reconstruction approach, we collect a large dataset for deformable object tracking (DOT). Our dataset contains $\sim$200 deformable motions of four types of objects: rope, paper, cloth, and hands (see examples in Fig.~\ref{fig:sample_data}). Their original appearance has various levels of textures, ranging from repetitive patterns to little or even no texture. For each motion, we provide 2D videos with and without markers from multiple viewpoints, 3D models of the deformed object, and tracked ground truth correspondences in both 2D and 3D. In total, our dataset has around one million video frames. Experimental results show DOT poses challenges to deformable surface tracking and reconstruction methods. Whereas by training on DOT, network performance significantly improved on weakly textured scenes, as being demonstrated on both our dataset and another deformable surface dataset (DeSurT~\cite{Wang_2019_ICCV}). 

In sum, our main contributions are as follows:  

\begin{itemize}

\item A novel type of invisible marker and an imaging system that allows simultaneous video acquisition of deformable objects with and without markers.
\item A template-based algorithm for 3D reconstruction and transferring correspondences from marker view to marker-free views. 
\item A large dataset, DOT, for deformable object tracking with ground truth tracked 2D and 3D correspondences.

\end{itemize}

\section{Related Work}\label{sec:related}

\noindent\textbf{Invisible light imaging.} 
The idea of invisible light imaging (e.g., in infrared $>750~\mathrm{nm}$ or ultraviolet $<380~\mathrm{nm}$ spectrum) has been widely explored in computational imaging. 
Many combine color images with near-infrared (NIR) images for denoising~\cite{dark_flash,9578549}, deblurring~\cite{7130624,7953319}, super-resolution \cite{7168367,8451274} and geometry estimation~\cite{6909896,7780638,Xia_2021_ICCV}. Notably, 
Wang \etal~\cite{Wang_2008_eurographics} uses infrared illumination to relight faces, in order to reduce the effect of uneven face color in a video conference setting. Krishnan and Fergus \cite{dark_flash} propose the ``dark flash" that
uses near-infrared (NIR) and near-ultraviolet (NUV) flashlights to replace the dazzling conventional flash. 
In another vein, Blasinski and Farrell~\cite{blasinski2017cmf} propose using narrow-band multi-spectral flash for color balancing, and Choe \etal~\cite{7780638} derive an
NIR reflectance model and use NIR images to recover fine-scale surface geometry.

Unlike widely studied NIR imaging, ultraviolet (UV) imaging has received much less attention. 

\noindent\textbf{UV fluorescence imaging.} UV fluorescence occurs when a substance
absorbs short wavelength light (such as UV light), and re-emits light at a longer
wavelength in visible spectrum~\cite{fluorescence,6247685}. 
This phenomenon has a wide range of imaging applications, including forensics~\cite{Richards_2013_forensic},
biomedical imaging~\cite{KLIJN2021319,Gray:06}, and material analysis~\cite{Yokoi_1991_SPIE,Giovanni_2008}. 

In computer vision, the fluorescent reflectance has
been studied for shape reconstruction~\cite{Treibitz2012ShapeFF}, immersive range
scanning~\cite{Hullin2008TOG}, inter-reflection removal~\cite{Fu2012ECCV},
multi-spectral reflectance estimation~\cite{9014556}, and material
classification~\cite{asano2018ECCV}, to name a few. 

Many prior works analyze the spectral response of
fluorescent reflectance, leading to techniques for appearance separation~\cite{5585093,6365191}, 
fluorescent relighting~\cite{6751166,6619035}, and camera spectral sensitivity estimation~\cite{Han2012CVPR}. 

In contrast to existing works, we use UV fluorescent markers to enrich surface features, without changing the object's appearance under normal lighting. We also design an imaging system that simultaneously captures video with and without fluorescent markers via time multiplexing.\\

\noindent\textbf{Deformable object reconstruction \& tracking.} Deformable objects are of great interest in computer vision as they are ubiquitous in daily life and their motions are often complex. Various sensor configurations have been explored for capturing deformable objects, 
including the use of single camera~\cite{7328075,Ngo_2015_ICCV,Qi_2015_BMVC}, 
multiple cameras~\cite{Schacter2014JIRS,Bickel2007siggraph,Bradley:2008}, 
and color cameras in tandem with depth cameras~\cite{zollhfer2014sig,bozic2020deepdeform} and 
with event cameras~\cite{Nehvi2021DifferentiableES}. 
Popular methods utilize local appearance to find
feature correspondences and then match 2D image features to a 3D shape
template for surface reconstruction and tracking~\cite{6619043,7010934,7110605,7464873}. 
However, these methods are hampered when the surface has
sparse or repetitive textures or has no texture at all. In these cases, the number of feature points is too scarce to establish reliable matching correspondences.
When the deformable surface has some but sparse textures, dense pixel-level template matching
may be helpful~\cite{4587499,6130447, Qi_2015_BMVC, Ngo_2015_ICCV}. Nevertheless, none of these methods can handle surfaces
with no texture---for example, a piece of white paper.

In recent years, many learning-based methods have been proposed for tracking and reconstructing deformable surfaces from a single view \cite{8491013,cgf13125,Tsoli_2019_ICCV,Shimada_2019}. These methods demonstrate great success in handling challenging cases when an object has repetitive or little texture. A comprehensive dataset is critical for training the neural networks, but datasets on deformable objects are quite limited. Most existing datasets are specific to a certain kind of object (\eg, \cite{8491013} on clothing and \cite{Wang_2019_ICCV} on paper, \etc). Furthermore, none of the real-captured datasets provide ground truth surface correspondences. In contrast, our DOT dataset has $\sim$200 deformable motions of four types of objects (\ie, rope, paper, cloth, and hands). Besides 2D videos with and without markers, we also provide high-quality 3D models and ground truth surface correspondences in 2D and 3D.

\section{Our Technique}\label{sec:method}

In this section, we present our method that utilizes invisible markers for deformable object reconstruction and tracking. We first introduce the optical properties of our markers (Sec.~\ref{sec:marker}), and then describe our imaging system for capturing deformable motions with and without markers (Sec.~\ref{sec:system}). Lastly, we present our template-based algorithm for 3D reconstruction and tracking (Sec.~\ref{sec:track}).  

\subsection{Invisible Markers}
\label{sec:marker}

\begin{figure}[t]
\centering
\includegraphics[width=1\linewidth]{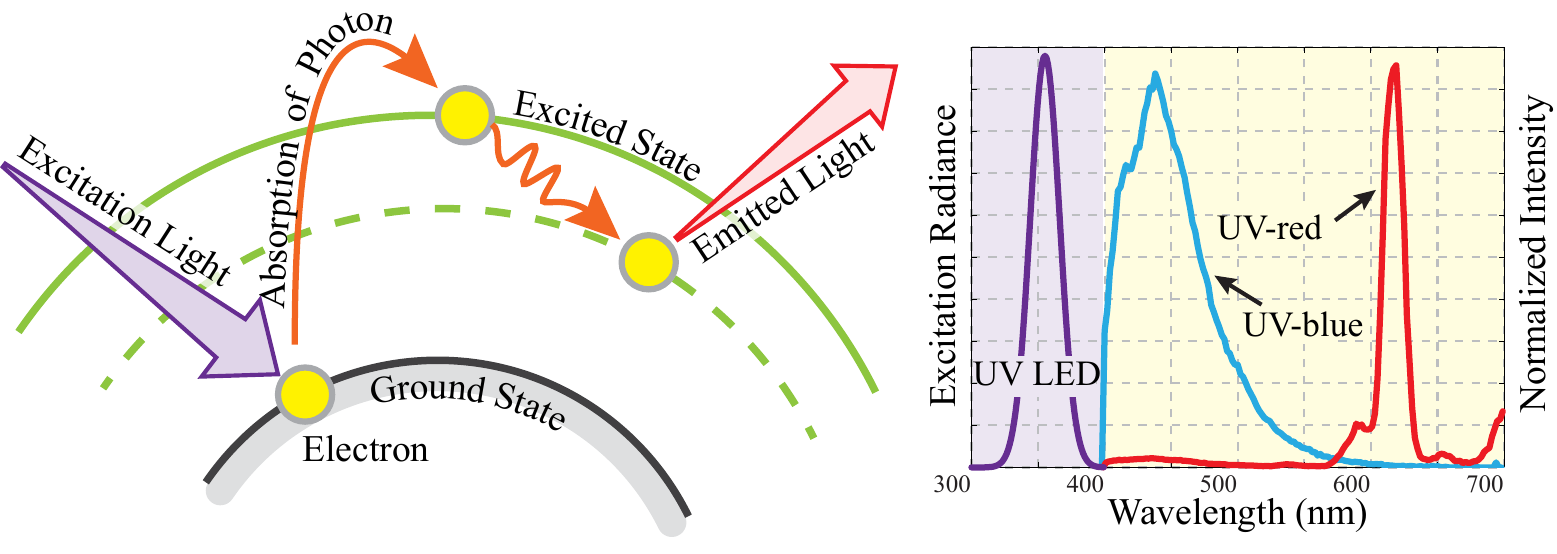}

\caption{\textbf{(left)} Illustration of the physics of fluorescence. 
    \textbf{(right)} The absorption and
    emission spectra of the fluorescent dyes used for our invisible markers. 
    We use two types of dyes with the peak emission in blue (UV-blue) and 
    red (UV-red). The purple curve shows the spectral profile
    of our UV light.}
\label{fig:uv_principle}
\end{figure}

We use UV fluorescent dyes to make markers invisible under normal light but visible under UV light. Fluorescent substance exhibits \emph{Stokes Shift}~\cite{stokes_shift}---an optical phenomenon wherein the material absorbs short wavelength light but
re-emits light at a longer wavelength. This phenomenon is caused by the material molecules' quantum behavior: when electrons of fluorescent material are irradiated by short wavelength light (\eg, UV light), they enter into an excited state after absorbing the light energy and then immediately de-excite and emit outgoing light at a longer wavelength (in the visible-light spectrum). 
This principle is illustrated in Fig.~\ref{fig:uv_principle}.

Specifically, we make a fluorescent dye solution 
and use it as the ink in a fountain pen to draw dot- or line-shaped markers on the object's surface.
The resulting markers are invisible under the
visible light, thereby preserving the object's original appearance.
The markers emit visible light (and thus become visible) only under UV light.
Moreover, the emitted light fades immediately---often within $10^{-8}\mathrm{sec}$---after the UV light is off.
Therefore, by using the UV light to trigger the markers' emission and 
synchronizing the trigger with the camera shutters, we can capture images
with and without the markers visible in a time-multiplexed manner
(see details in Section~\ref{sec:system}).

\begin{figure}[t]
\centering
\includegraphics[width=0.99\linewidth]{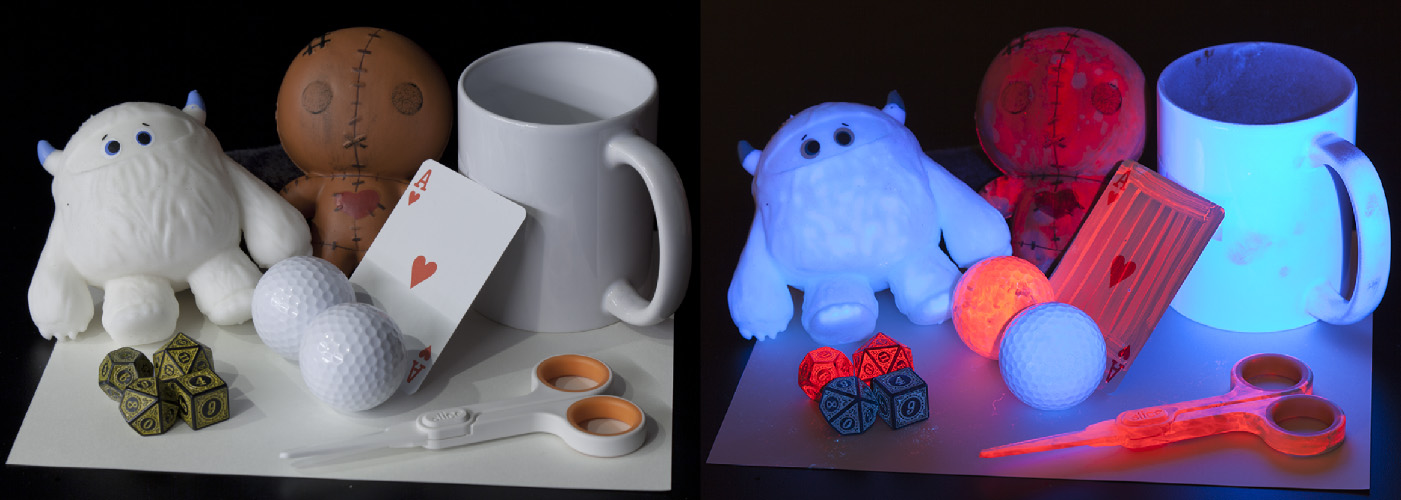}

\caption{Objects covered with our fluorescent ink (UV-blue and UV-red). 
    \textbf{(left)} Image under visible light. 
    \textbf{(right)} Image under UV light, where the fluorescent colors become visible.}
\label{fig:uv_object}
\end{figure}

We look for fluorescent dyes that satisfy two criteria: 
1) the fluorescent emission under UV light has
high contrast and strong visibility; and
2) the dyes are biologically safe and non-toxic to human skin. In our experiments, we use the MaxMax UV
dyes\footnote{https://maxmax.com/phosphorsdyesandinks}. For scenes with multiple objects (\eg, hand-object interaction), we use two types of fluorescent dyes: one emits blue light when excited and the other red (they are referred to as \emph{UV-blue} and \emph{UV-red}). In this way, multiple objects in one image can be easily separated by color (see Fig.~\ref{fig:uv_object}). 


\textbf{Fluorescence Detection.} Since fluorescent emissions are narrow-banded in wavelength
(see Fig.~\ref{fig:uv_principle}), their hue values in fluorescent images usually have small ranges.
Exploiting this fact, we can detect markers by using hue values:
we convert the images into the HSV color space, and label pixels whose hue values fall
into a certain fluorescent dye's emission range.
This marker detection in HSV color space is robust even when the surface itself is 
fluorescent. This is because the narrow-banded fluorescent emission peaks
at a specific spectral location, and it is very unlikely that the surface's
emission peaks at the same location as our two types of markers.

\textbf{Dye Preparation.}
To prepare the fluorescent ink for drawing, we dissolve the UV-blue dye
in 70\% alcohol and UV-red in acetone.
Both types of markers can be triggered for emission under $365~\mathrm{nm}$ UV light.
We chose this UV light spectrum, because it is in the range of UVA, the safest UV spectrum 
for human skin, abundant in natural sunlight.

Caution is needed when choosing the solution's concentration level.
The higher the concentration is, the brighter the fluorescent emission under the UV light becomes. 
On the one hand,
if the fluorescent emission is too bright, the markers will saturate 
image pixels, rendering the marker detection based on hue values much harder.
On the other hand, if the concentration is too low, the markers appear too dim,
and the detection also becomes less robust.

\begin{figure}[t]
\centering
    \includegraphics[width=0.99\linewidth]{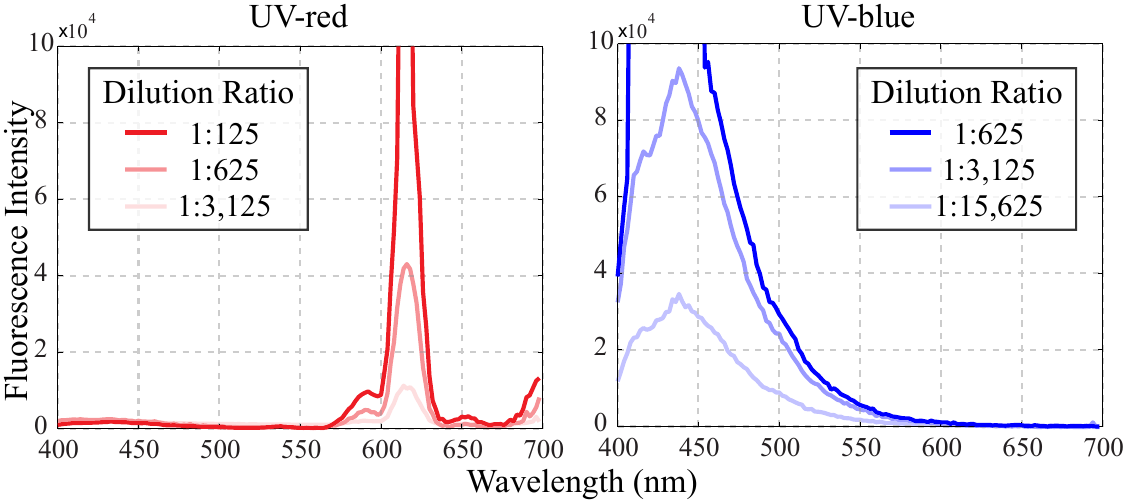}

    \caption{The spectral response of fluorescent solutions with different concentration levels
        of the UV-red (left) and UV-blue (right).
    \label{fig:curves}}
\end{figure}

We choose the concentration level through systematic measurements.
We test both UV-blue and UV-red dyes. For each type of dyes,
we prepare the fluorescent solution of different concentrations
spaced by $1/5$ dilution ratio (i.e., $1/5$, $1/25$, $1/125$, ...), 
covering concentration levels from $1/5$ to $1/15625$.
We then measure the spectral responses of each sample using a modular multimode
microplate reader (BioTek Synergy H1).  
Under $365~\mathrm{nm}$ UV light, the measurement records the response of 
fluorescent emission in the range of $400~\mathrm{nm}$ to
$700~\mathrm{nm}$ (visible light spectrum) with a $2~\mathrm{nm}$ step.
A subset of the measured response curves are shown in Fig.~\ref{fig:curves}.
Through the measurements, we find that 
for both types of dyes, the desirable concentration level is $1/625$, 
and we thereafter use this ratio in experiments.

Please see the supplementary material for more details about our fluorescent dyes, including their spectral responses on different types of materials, multi-object separation example, and safety measures.

\subsection{Imaging System} 
\label{sec:system}

We design and build a novel imaging system for capturing videos of deformable objects with and without markers, by leveraging UV fluorescence. 

\begin{figure}[t]
\centering
\includegraphics[width=1.0\linewidth]{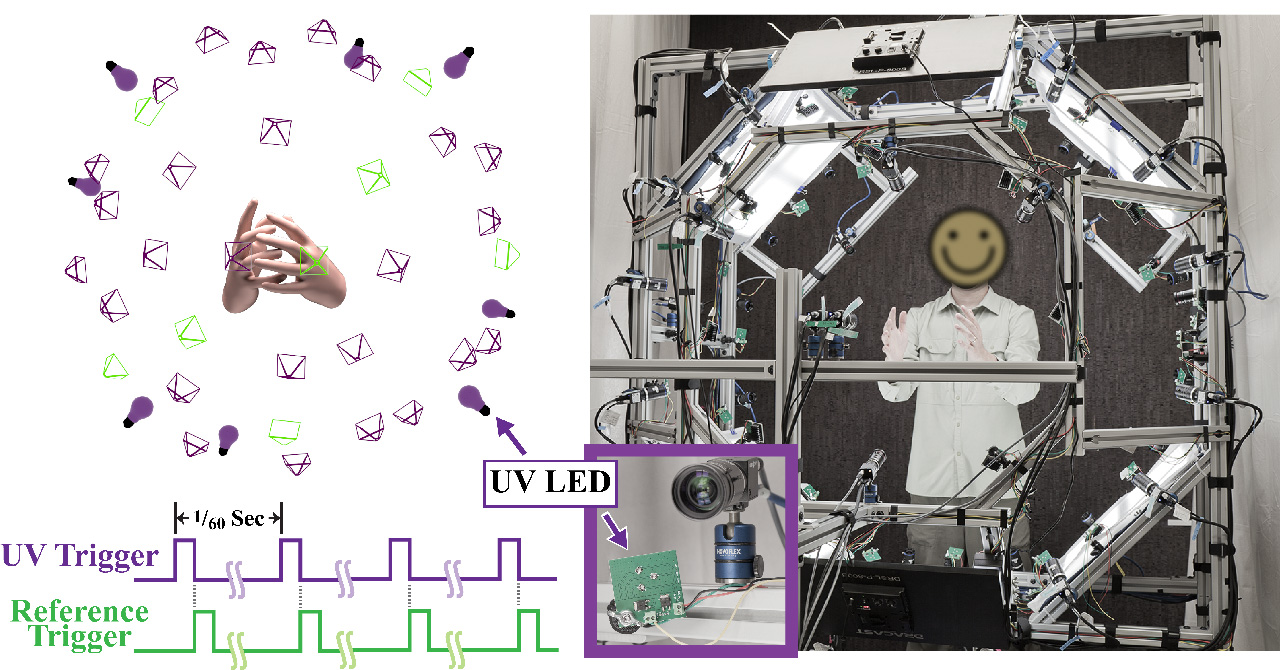}
\caption{\textbf{(left)} Conceptual illustration of our system with trigger scheme (green color refers to reference cameras, and purple color refers to UV cameras). \textbf{(right)} The real physical setup of our system with a zoom-in view of the UV LED unit.}
\label{fig:system_setup}
\end{figure}

\textbf{System configuration.} Our imaging system consists of 42 global-shutter color cameras each with a $16~\mathrm{mm}$ lens and 60 UV LED light units. All cameras and lights are uniformly mounted on a rhombicuboctahedral rig frame, facing inward to the frame center. Fig.~\ref{fig:system_setup} shows a conceptual illustration and the real physical setup of our acquisition system. Since the rhombicuboctahedral frame is a discrete approximation of a sphere, distances from the cameras and lights to the center of the frame (where the deformable object is located) are about the same ($\sim 75~\mathrm{cm}$), which avoids uneven light attenuation. Please see the supplementary material for more details about the specs of cameras and light sources, as well as the camera calibration procedure.

\textbf{Trigger scheme.} Videos of the deformable object with and without markers are captured in a time-multiplexed fashion. To this end, we group the cameras into two sets:
one set is triggered when the UV lights are turned on (referred to as \emph{UV cameras\footnote{These cameras are still regular cameras sensitive to visible light. Here ``UV" means that they capture images under UV lights.}}), and the other set is triggered when the UV lights are off (referred to as \emph{reference cameras}). In this way, the UV cameras capture the object with markers, while the reference cameras capture its original untouched appearance. In practice, we use 33 UV cameras and 9 reference cameras for high-quality tracking and 3D reconstruction.

All cameras capture videos at a frame rate of $60~\mathrm{fps}$. The time interval between two consecutive frames is therefore $16~\mathrm{ms}$. Since this interval is much larger than the camera's exposure time, we can arrange the exposure period of the two sets of cameras within the $16~\mathrm{ms}$ window back-to-back with no overlap. Our triggering scheme is illustrated in Fig.~\ref{fig:system_setup}. We custom-build the FPGA control board for syncing and triggering the cameras and lights.  

In practice, we use $2~\mathrm{ms}$ exposure time for all cameras. The delay between the videos with and without markers is therefore $2~\mathrm{ms}$. Although the time difference is very short, we still use interpolation to reduce the amount of possible misalignment. The influence of the delay is studied in Sec.~\ref{sec:ablation}.

\subsection{Reconstruction and Tracking}
\label{sec:track}
\begin{figure}[t]
\centering
\includegraphics[width=1\linewidth]{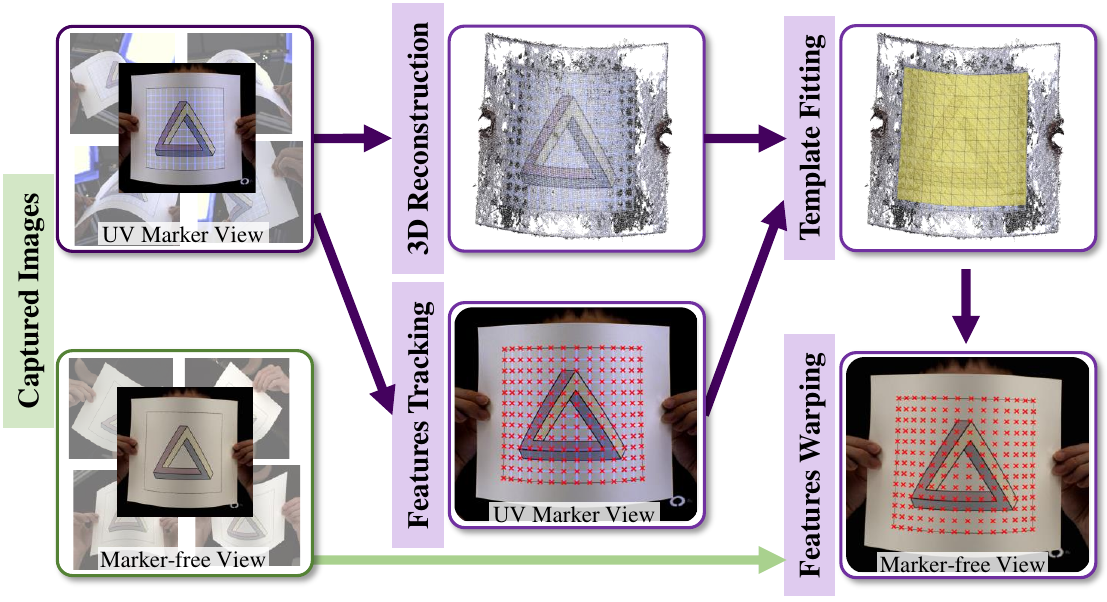}
\caption{Warping features from marker view (under UV lighting) to maker-free view (under normal lighting).}
\label{fig:pipeline}
\end{figure}

Given the videos with and without markers, we first detect surface correspondences and perform template-based 3D reconstruction using the video frames that capture markers. We then apply the tracked correspondences to marker-free videos as ground-truth labels. Fig.~\ref{fig:pipeline} shows our pipeline.

\textbf{2D marker detection \& 3D reconstruction.}
To detect markers in UV fluorescent images (captured under UV lighting, in which markers appear), we convert the images into the HSV color
space. The fluorescent reflectance typically exhibits high
Saturation (S) and Value (V) values, and their Hue (H) values fall into a small range depending on the dye's emission profile (\eg, $\mathrm{H}\in[0,15]$ for UV-red and $\mathrm{H}\in[110,125]$ for UV-blue). We therefore threshold the H value to detect marker pixels. These markers are used as features for 3D reconstruction and temporal tracking. 

Any 3D photogrammetry-based reconstruction algorithm (\eg, COLMAP \cite{schoenberger2016sfm, schoenberger2016mvs}, Meshroom \cite{alicevision2021}) can be used on our multi-view marker images to obtain a 3D point cloud for each frame of a motion sequence.
In our experiment, we first compute dense disparity maps through semi-global
matching~\cite{hirschmuller2005accurate}, and then project the depth maps to point cloud. We also perform cross-view validation to enhance the reconstruction accuracy.

\textbf{Template fitting.}
Next, we fit each 3D point cloud to an object-dependent predefined 3D template and project the 2D feature correspondences onto the 3D model. We use different templates for different types of object (see details about the templates in Sec.~\ref{sec:dataset}). 
Here we present our fitting algorithm in general that is applicable to all data. 

Consider a 3D point cloud $\mathcal{S}$ consisting of $M$ points $\textbf{p} =
\{\bm{p}_1,\bm{p}_2,...,\bm{p}_M\}$ and a 3D shape template described by $N$ vertices $\textbf{v}
= \{\bm{v}_1, \bm{v}_2, ... ,\bm{v}_N\}$ and $K$ faces.
Our goal is to deform the shape template so that it aligns closely to the 3D point cloud.

We adopt the embedded deformation graph~\cite{sumner2007embedded} to deform the template:
for every vertex $i$ on the template shape, its deformed position is described 
by $\bm{v}_i + \bm{t}_i$. To ensure deformation smoothness, every vertex $i$ also has a local region of influence. 
Its influence is described by a rotational matrix $\bm{R}_i\in \mathrm{SO}(3)$, which maps any point 
$\bm{p}$ in its local region to the position $\bm{p}'$ according to
\begin{equation}
\footnotesize
    \bm{p}' = \bm{R}_i(\bm{p} - \bm{v}_i) + \bm{v}_i + \bm{t}_i.
\end{equation}
We determine $\bm{t}_i$ and $\bm{R}_i$ (for $i=1..N$) by solving the following optimization problem:
\begin{equation}\label{eq:opt}
\footnotesize
    E_{\text{total}} = E_{\text{fit}} + \lambda_{\text{m}}E_{\text{marker}} + \lambda_{\text{s}}E_{\text{smooth}}.
\end{equation}

The first term $E_{\text{fit}}$  measures the $\ell_2$ distance
between the 3D point cloud and the deformed template mesh in two ways:
for each point $j$ in the point cloud $\mathcal{S}$, its distance to the closest 
vertex and the closest face on the template.  
$E_{\text{fit}}$ is thus a summation of two terms, namely,
\begin{equation}
\footnotesize
    \begin{split}
    E_{\text{fit}} = & \sum_{j=1}^{M}\underbrace{\lVert\bm{v}_{c(i)}+\mathbf{t}_{c(j)}-\bm{p}_j\rVert^{2}_2}_{\text{point-to-vertex distance}} + \\
    & \beta \sum_{j=1}^{M}\underbrace{\lVert \bm{n}_{c(i)}^{\top}(\bm{v}_{c(j)}+\mathbf{t}_{c(i)}-\bm{p}_j)\rVert^2_2}_{\text{point-to-face distance}},
    \end{split}
\end{equation}
where $c(j)$ indicates the index of the deformed template vertex closet to the point $\bm{p}_j$,
$\bm{n}_{c(j)}$ denotes the vertex normal, and 
$\beta$ is a weight for balancing the two terms.

The second term $E_{\text{marker}}$ measures the distance between the detected
2D markers and the projected marker positions from the template mesh.
Consider $N_f$ markers. Their 3D positions on the undeformed template mesh, denoted by
$\bm{x}_j$ for $j=1...N_f$, are initialized at the beginning of the capture session. $\bm{x}_j$ can be expressed using 
the barycentric coordinate $\bm{\alpha}_j = [\alpha_{j,1}\; \alpha_{j,2}\; \alpha_{j,3}]$ on the template triangle where it is located, that is,
$\bm{x}_j = \sum_{k=1}^3\alpha_{j,k} \bm{v}_{j,k}$,
where $\bm{v}_{j,k}$ ($k=1,2,3$) are the vertex positions of the template triangle.
With these notations, $E_{\text{marker}}$ is defined as
\begin{equation} \label{eq:marker}
\footnotesize
    E_{\text{marker}} = \sum_{i=1}^{N_v} \sum_{j=1}^{N_f} w_{ij} \lVert \bm{p}_{ij} - \bm{\pi}_i\bm{\tilde{x}}_j \rVert^{2}_2,
\end{equation}
where $\bm{\tilde{x}}_j$ is the $j$-th marker's 3D position on the deformed template mesh
(i.e., $\bm{x}_j = \sum_{k=1}^3\alpha_{j,k} (\bm{v}_{j,k}+\bm{t}_{j,k})$),
$N_v$ is the number of UV camera views; 
$w_{ij}$ is the confidence weight for the marker $j$ being viewed from the $i$-th camera. 
If the marker $j$ is occluded from the $i$-th camera, $w_{ij}$ vanishes. 
Moreover, $\bm{p}_{ij}$ is the 2D position of a marker $j$ (if not occluded) on the $i$-th camera view,
and $\bm{\pi}_i$ is the projection matrix of the $i$-th camera.

The last term $E_{\text{smooth}}$ regulates the smoothness of the template mesh's deformation.
This is where the local influence of each vertex $i$ (and hence $\bm{R}_i$) is involved.
Following a similar term defined in~\cite{sumner2007embedded} (called $E_{\text{reg}}$ therein), 
$E_{\text{smooth}}$ encourages the mesh deformation to be locally rigid, defined as
\begin{equation}
\footnotesize
    E_{\text{smooth}} = \sum_{j=1}^{N}\sum_{k\in\mathcal{N}(j)} \gamma_{jk} 
    \lVert \bm{R}_{j}\bm{v}_{kj}- \bm{v}_{kj} + \bm{t}_{j} -\bm{t}_{k}\rVert^{2}_2,
\end{equation}
where $\bm{v}_{kj}$ is a shorthand for $\bm{v}_{kj} = \bm{v}_k - \bm{v}_j$;
$\mathcal{N}(j)$ is the neighboring vertices of vertex $i$ (here defined as 
the 10-nearest neighbors of $i$); and $\gamma_{jk}$ is a
weight parameter determined by distance between $\mathbf{v}_j$ and $\mathbf{v}_k$. 

When solving the optimization problem~\eqref{eq:opt}, we express each $\bm{R}_i$ in 
Lie algebra $\mathrm{SO}(3)$ and use the Levenberg-Marquardt optimizer.
The optimization starts with large $\lambda_{\text{s}}$ and
$\lambda_{\text{m}}$ values, and gradually reduces them in iterations 
until the optimization converges.

\textbf{Feature warping.} 
Finally, we project 3D marker features back to marker-free reference views, in order to label ground truth surface correspondences on the object's original appearance. Note that the features cannot be differently warped from marker view to marker-free view in 2D as their depths are not known. This allows us to pair deformable objects with little or no textures with their ground truth surface correspondences. 

Recall that our videos with and without markers have a few millisecond delays. In cases when the motion is slow, this delay is small enough to be ignored. However, when an object moves too fast, the delay may cause a noticeable misalignment between the projected features and the reference frame. We alleviate the misalignment by linearly interpolating the 3D models and feature points of two consecutive frames to the time instant at which the reference frame is captured (see details of the interpolation algorithm in supplementary material). This strategy, albeit simple, is effective in reducing the misalignment.

\section{Deformable Object Tracking Dataset}
\label{sec:dataset}

\begin{figure}[t]
\centering
\includegraphics[width=1.0\linewidth]{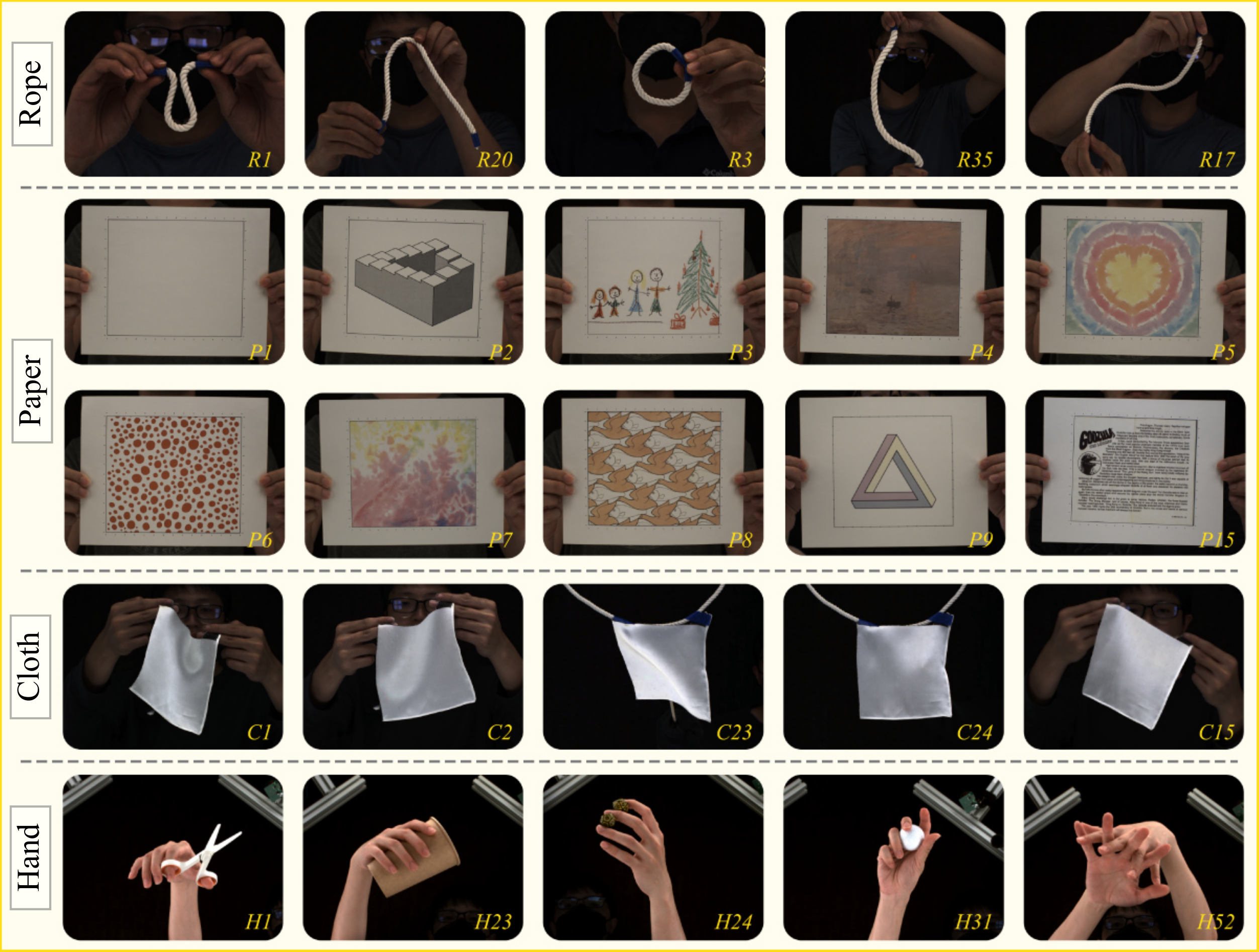}
\caption{Sample scenes in our DOT dataset. Here we show marker-free images under normal lighting.}
\label{fig:dataset}
\end{figure}

Using our system, we collect a large dataset for deformable object tracking, which we refer to as DOT. Our dataset contains deformable motions of four types of objects: rope, paper, cloth, and hand. Original appearance of these objects has different levels of textures, ranging from repetitive texture to little or no texture (see examples in Fig.~\ref{fig:dataset}). We draw different fluorescent patterns on the objects, introducing rich features for correspondence matching under UV lighting. In total, we have $\sim 200$ deformable motion sequences. Each sequence has multi-view videos with and without markers, per-frame 3D models and point clouds, and ground truth correspondences in both 2D and 3D. The total number of maker-free video frames is around one million. The details of our DOT dataset are summarized in Table~\ref{table:dot} (see the supplementary material for video data samples of DOT). 

\begin{table}[b]
\centering
\footnotesize
\caption{\label{table:dot} Summary of DOT per category.}
    \begin{tabular}{||c c c c c||} 
    \hline
    Category & Rope & Paper & Cloth & Hand \\  [0.5ex] 
    \hline\hline
    \# of Motion Seq. & 60 & 20 & 30 & 90\\
    Seq. Length (time) & 10s & 30s & 10s & 5s \\
    Seq. Length (frame) & 600 & 1800 & 600 & 300 \\
    \# of Viewpoints & 10 & 6 & 6 & 10 \\
    Total \# of Frame & 360K & 216K & 108K & 270K \\  
    Template Type & Joint & Grid & Grid & Mesh  \\[1ex] 
    \hline
    \end{tabular}
\end{table}

Our dataset is versatile for network training due to its data abundancy and diversity. By pairing the marker-free videos with ground-truth correspondences, we can train networks for deformable object tracking. In addition, we provide videos from different viewpoints that can suit the need for different imaging configurations. By pairing either single-view or multi-view marker-free frames with their 3D models, our dataset can be used for training deformable object reconstruction networks. As we have an ample amount of objects with little or no textures, training on our dataset can improve networks' performance in handling these challenging cases. We will release DOT upon publication.
In the following, we discuss data preparation in details for each category.

\textbf{Rope Scenes.} We capture ropes with different lengths (5$^{\prime\prime}$, 10$^{\prime\prime}$, and 12$^{\prime\prime}$) and thicknesses (1/2$^{\prime\prime}$, 1/4$^{\prime\prime}$, and 1/8$^{\prime\prime}$). The motions we perform include swinging, twisting, shaking, pulling, and stretching. We treat the rope as a 1D object and use connected joints as its template. We attached blue tapes at the two ends of the rope to indicate the start- and end-point of our rope object. As for invisible fluorescent markers, we use UV-red to draw dots on the rope with 0.5$^{\prime\prime}$ intervals. Depending on the length of the rope, we use different numbers of joints in the template (\ie, 10, 16, and 18 nodes), and these joints are mapped to the invisible markers as correspondences.

\textbf{Paper Scenes.} We use letter-sized non-fluorescent papers, and print different patterns on them to create variations in texture. These variations include rich texture (\eg, texts and random dots), repetitive texture, smooth texture (\eg, tie-dye patterns), weak texture (\eg, sparse line drawings), and no texture (pure white paper). Note that these printed patterns are all visible and are considered as the paper's original appearance. To introduce invisible marker pattern, we draw a 13$\times$15 line grid with UV-blue for all scenes. We use the same grid as their 3D template. The grid intersections are used as correspondences for tracking and reconstruction. We record the deformable motions by hand-twisting the paper.

\textbf{Cloth Scenes.} We use white silk cloth of different sizes. All motion sequences in the cloth scenes are fully textureless. Same as the paper scenes, we draw a 2D line grid as invisible patterns on the cloth. The size of the grid is determined by the cloth size. As cloth is more deformable than paper, we use a rigid frame to stretch the cloth when drawing the grid pattern. Our deformable motions include swinging, shaking, blowing, and stretching. The motions are induced by hand manipulation or wind blowing. We also perform motions at different speeds. 

\textbf{Hand Scenes.} We include a variety of common hand motions and gestures performed by a single hand or two hands. We also include interactive motions between hand(s) and objects, such as scissors, mugs, dice, and toys. For invisible markers, We use UV-blue to draw random dots on hands (see supplementary material for discussions on dye safety on human skin). For hand-object interactions, we use UV-red to fully cover the object, such that we can use the hue difference to separate hand points and object points. This largely improves the accuracy of template fitting (see the result in the supplementary material). For reconstruction and tracking, we use the MANO \cite{MANO:SIGGRAPHASIA:2017} model as a 3D template for hand. The invisible markers are registered to the template using the rest pose. In our acquisition, all hand motions start with the rest pose for feature initialization. Since we have 33 UV cameras for capturing images with markers, our 3D reconstruction and tracked correspondences are very accurate and robust, even in case of heavy occlusion (\eg, crossed hands, or hand occluded by object).

\section{Experiments}
\label{sec:exp}

We first perform experiments to evaluate the performance of our imaging system and effectiveness of tracked markers (Sec.~\ref{sec:ablation}). We then benchmark state-of-the-art algorithms on tracking and reconstruction on our DOT dataset (Sec.~\ref{sec:benchmark}). We also demonstrate the benefit of using DOT for network training (Sec.~\ref{sec:finetune}).

\begin{figure}[t]
\centering
\includegraphics[width=1.0\linewidth]{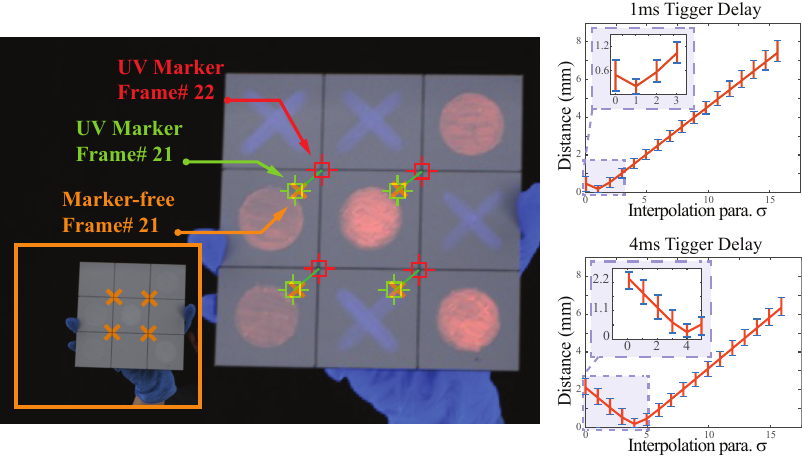}
\caption{Influence of trigger delay. \textbf{(left)} Sample images of our grid target. We can see apparent motion between neighboring frames (green circles in \#21 vs. red circles in \#22). For the same frame (\#21), features between the marker-free view (orange crosses) and marker view (green circles) are slightly misaligned due to trigger delay. \textbf{(right)} Point-to-ray distance curves with respect to the interpolation parameter $\sigma$ for $1~\mathrm{ms}$ and $4~\mathrm{ms}$ delays. } 
\label{fig:intr_diff}
\end{figure}

\subsection{System Performance}
\label{sec:ablation}

\noindent\textbf{Influence of delayed trigger.}  We first study the influence of delayed trigger: how much misalignment between the reference and UV views would be caused by the trigger delay, and how effective our interpolation algorithm is at reducing the misalignment. This analysis is very important as our goal is to leverage the ``invisible" makers in the UV view for reconstruction and tracking under the reference view (without markers), and we want to make sure that the features detected and tracked in frames with markers are well synchronized with marker-free frames.

We perform experiments using a board paper with a grid. The grid can be seen in both reference and UV views. We use the grid corners as features for measuring the pixel shift caused by the trigger delay. Sample images of the target are shown in Fig.~\ref{fig:intr_diff}. We illustrate feature points from corresponding reference and UV views (marked in orange and green, respectively), as well as those from the consecutive UV view (marked in red). We can see that features from the two consecutive UV images have apparent shifts (whose time interval is $16~\mathrm{ms}$). The feature misalignment between the corresponding UV and reference views is slight in this case but could vary depending on the object's motion speed (\eg, the faster the speed, the larger the shift).

We then quantitatively measure the amount of misalignment for different trigger delays, and evaluate the effectiveness of the following interpolation scheme. We first compute the 3D coordinates of the feature points via ray triangulation from all UV views. As the reference views are too sparse for accurate triangulation, we only trace out rays from features on the reference view. We use $\mathbf{r}^t$ to denote the rays traced out from the reference view at time $t$. Assume the 3D points computed by the UV views at time $t$ and time $t+1$ are $\mathbf{v}^t$ and $\mathbf{v}^{t+1}$, respectively, we then linearly interpolate $\mathbf{v}^t$ and $\mathbf{v}^{t+1}$ to obtain an intermediate point $\mathbf{\bar{v}}^{\sigma}$. We calculate the point-to-ray distance from $\mathbf{\bar{v}}^{\sigma}$ to $\mathbf{r}^t$ and use it to measure how well the features are aligned in 3D. 

We test on two delay values: $1~\mathrm{ms}$ and $4~\mathrm{ms}$. We compute the point-to-ray distance with respect to various interpolation parameters $\sigma$, from $0$ to $15$. When $\sigma = 0$, it is equivalent as directly using $\mathbf{v}^t$ (\ie, no interpolation). We plot the curves of point-to-ray distance with respect to different $\sigma$ in Fig.~\ref{fig:intr_diff}. We can see that our interpolation is particularly useful when the trigger delay is large. For example, in the case of $4~\mathrm{ms}$ delay, the misalignment is $2.23~\mathrm{mm}$ without interpolation. Our interpolation brings down the error to $0.28~\mathrm{mm}$ (when $\sigma = 4$).

\begin{figure}[t]
\centering
\includegraphics[width=1.0\linewidth]{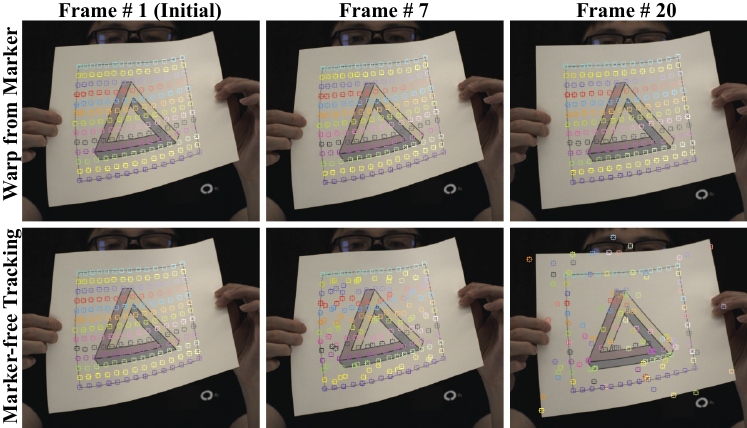}
\caption{Feature tracking results with vs. without using fluorescent markers (paper scene $P9$).}
\label{fig:track_deformable_compare}
\end{figure}

\noindent\textbf{Tracking results with vs. without markers.} In order to show the effectiveness of our fluorescent markers on lack-of-texture surfaces, we compare the correspondence tracking results by using our markers versus without using the markers (\ie, directly apply the tracking algorithm on the marker-free reference view). 

Fig.~\ref{fig:track_deformable_compare} shows comparison results on a paper scene ($P9$). We use the tracking algorithm provided in the commercial software R3DS Wrap4D2. Features are initialized as grid-line corners. We can see that our tracked correspondences computed using the marker view are reliable and robust over time on this paper scene, which lacks texture in large regions. In contrast, if we only use the marker-free video, the algorithm will lose track of most of the feature points within 20 frames (the entire video has 1800 frames). 
Therefore, with our fluorescent markers, the tracking result is more accurate and robust, with the mismatch rate largely reduced.

\subsection{Benchmark Experiments}
\label{sec:benchmark}
We benchmark state-of-the-art methods on optical flow, hand reconstruction, and deformable object tracking and reconstruction using our DOT dataset. 

\noindent\textbf{Optical Flow Methods.}

We test on five state-of-the-art optical flow algorithms: classical variational optical flow as implemented in OpenCV~\cite{farneback2003two} (referred to as ``GF"), the two most popular learning-based optical flow networks---PWC-Net~\cite{sun2018pwc} and RAFT~\cite{teed2020raft}, deformable local feature enhanced optical flow network---DALF~\cite{potje2023enhancing}, and globally consistent motion tracking algorithm---OmniMotion \cite{wang2023omnimotion}.

\begin{table}[t]
\centering
\footnotesize 
    \caption{Quantitative comparisons of flow errors on rope, cloth and paper scenes. All errors are reported in pixel unit.}    
    \label{tab:benchmark} 
    \setlength\tabcolsep{4.5pt}
    \begin{tabular}{cccccccc} 
    \hline
        \multirow{2}{*}{Method} & \multicolumn{7}{c}{Flow Error (\textbf{Rope})} \\
                              &  $R14$ &  $R15$ &  $R16$ &  $R17$ &  $R18$ &  $R19$ & Avg. \\ 
        \hline
        GF\cite{farneback2003two} 
                              & 0.934 & 0.529 & 0.882 & 0.227 & 1.990 & 0.486 & 0.841 \\\cline{2-8}
        PWC\cite{sun2018pwc} 
                              & 0.764 & 0.525 & 0.779 & 0.834 & 1.941 & 0.703 & 0.929 \\\cline{2-8}
        RAFT\cite{teed2020raft}
                              & 1.172 & 0.887 & 1.052 & 0.984 & 2.360 & 1.176 & 1.272 \\\cline{2-8}
        OMNI\cite{wang2023omnimotion} 
                              & 1.448 & 0.871 & 2.138 & 0.701 & 2.971 & 1.166 & 1.549 \\\cline{2-8}
        DALF\cite{potje2023enhancing}
                              & 0.793 & 0.602 & 0.884 & 0.731 & 1.526 & 0.645 & 0.864 \\
        \hline
     \hline
        \multirow{2}{*}{Method} & \multicolumn{7}{c}{Flow Error (\textbf{Cloth})} \\
                              &  $C2$ &  $C3$ &  $C4$ &  $C5$ &  $C9$ &  $C10$ & Avg. \\ 
        \hline
        GF\cite{farneback2003two}
                              & 1.587 & 0.678 & 2.231 & 2.970 & 2.923 & 2.935 & 2.221 \\\cline{2-8}
        PWC\cite{sun2018pwc}
                              & 0.907 & 0.689 & 1.190 & 1.210 & 1.171 & 1.230 & 1.066 \\\cline{2-8}
        RAFT\cite{teed2020raft}
                              & 1.006 & 0.689 & 1.057 & 1.298 & 1.546 & 1.365 & 1.160 \\\cline{2-8}
        OMNI\cite{wang2023omnimotion} 
                              & 1.001 & 0.679 & 1.097 & 1.380 & 1.423 & 1.471 & 1.175 \\\cline{2-8}
        DALF\cite{potje2023enhancing}
                              & 4.912 & 2.165 & 5.920 & 6.339 & 9.627 & 6.048 & 5.835 \\
        \hline
       \hline
        \multirow{2}{*}{Method} & \multicolumn{7}{c}{Flow Error* (\textbf{Paper})} \\
                              &  $P1$ &  $P2$ &  $P4$ &  $P5$ &  $P7$ &  $P8$ & Avg. \\ 
        \hline
        {GF\cite{farneback2003two}} 
                              & 5.429 & 4.106 & 6.069 & 4.747 & 5.034 & 3.354 & 4.662 \\\cline{2-8}
        {PWC\cite{sun2018pwc}} 
                              & 8.215 & 0.564 & 0.394 & 0.402 & 0.362 & 0.484 & 0.441 \\\cline{2-8}
        {RAFT\cite{teed2020raft}} 
                              & 196.2 & 0.625 & 0.408 & 0.405 & 0.374 & 0.435 & 0.449 \\\cline{2-8}
        {OMNI\cite{wang2023omnimotion}}  
                              & 1.360 & 0.954 & 0.852 & 0.842 & 0.860 & 0.885 & 0.879 \\\cline{2-8}
        {DALF\cite{potje2023enhancing}}  
                              & 16.91 & 5.678 & 1.833 & 0.745 & 0.966 & 1.061 & 2.057 \\
        \hline
    \end{tabular}
    
      \scriptsize{*The average errors of paper scenes are calculated without using the $P1$ errors.}
\end{table}

We perform experiments on rope, cloth, and paper scenes. For the sake of space, here we mainly show quantitative evaluation results on a subset of scenes in each category. Rope scenes all have the same thread texture, but the ropes have different lengths and thicknesses. Cloth scenes are all pure white silk cloth with various motions. Paper scenes have different variations of textures: $P1$ has no texture (\ie, white paper); $P2$ has weak textures and large blank regions; $P4$, $P5$, and $P7$ has smooth textures with no sharp edges; $P8$ has a repetitive pattern (see Fig. \ref{fig:dataset} for these scenes). 

For rope scene, we only use the flow vectors of their 1D node joints. For both paper and cloth scenes, we interpolate per-pixel dense flow from the optical flows at grid corners. Specifically, our dataset provides ground-truth 2D and 3D coordinates for their grid templates, from which we can compute the ground-truth 2D and 3D optical flows at grid corners. We run these algorithms on the marker-free video captured from a frontal view. We compute the mean square errors between the estimated optical flow vectors and our ground truth ones. The errors are reported in Table~\ref{tab:benchmark} for all three categories. For each scene, the error is averaged on the entire motion sequence. The average error in the last column is computed using all available scenes. We can see our cloth scenes and the white paper scene ($P1$) pose challenges to all methods. Since $P1$ errors are too large, we exclude them for computing the average errors.

\begin{figure}[t]
\centering
\includegraphics[width=1.0\linewidth]{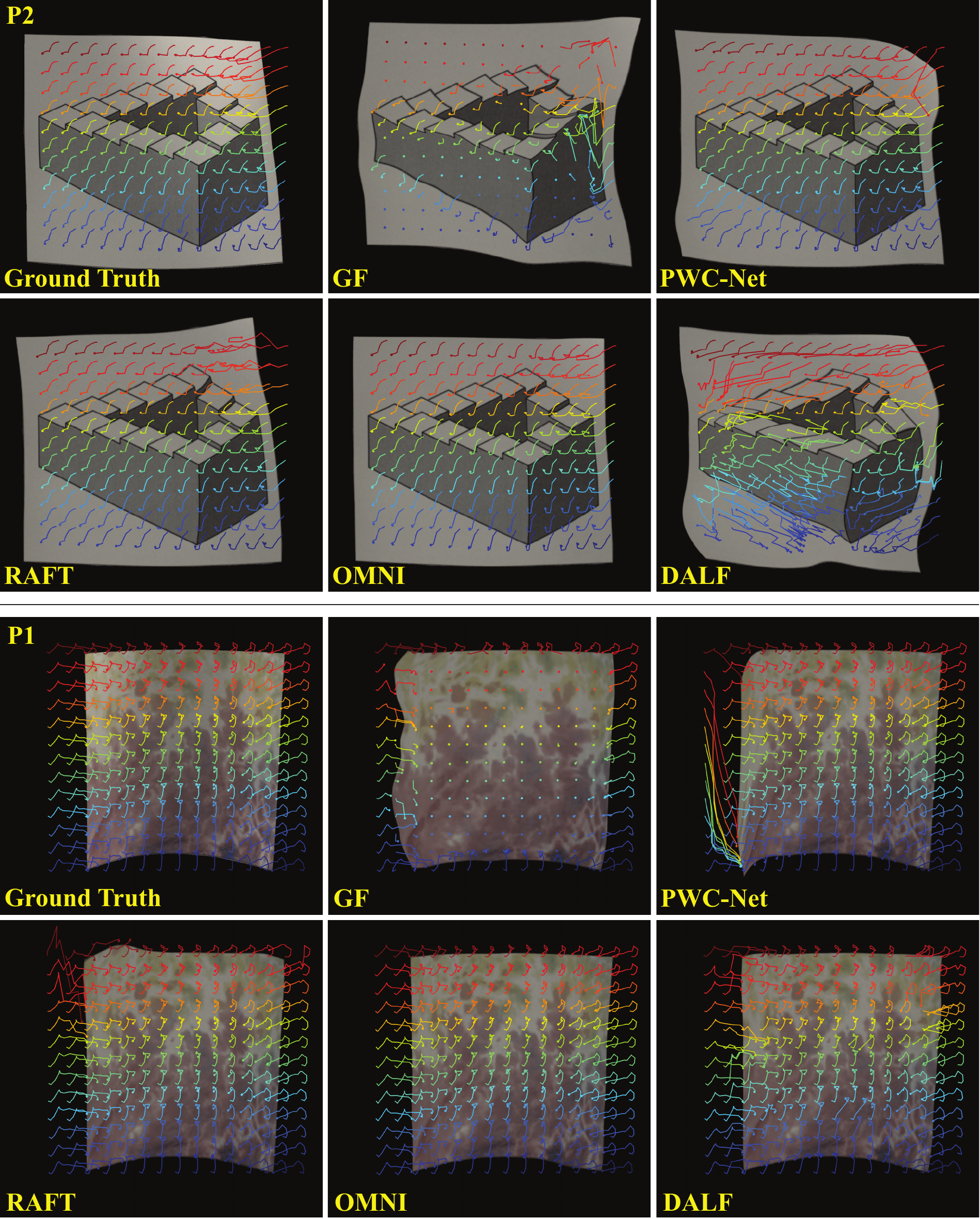}
\caption{Qualitative comparisons on warped images with accumulated flow trajectory (paper scene $P1$ and $P2$). }
\label{fig:compare_opt_paper}
\end{figure}

\begin{figure}[t]
\centering
\includegraphics[width=1.0\linewidth]{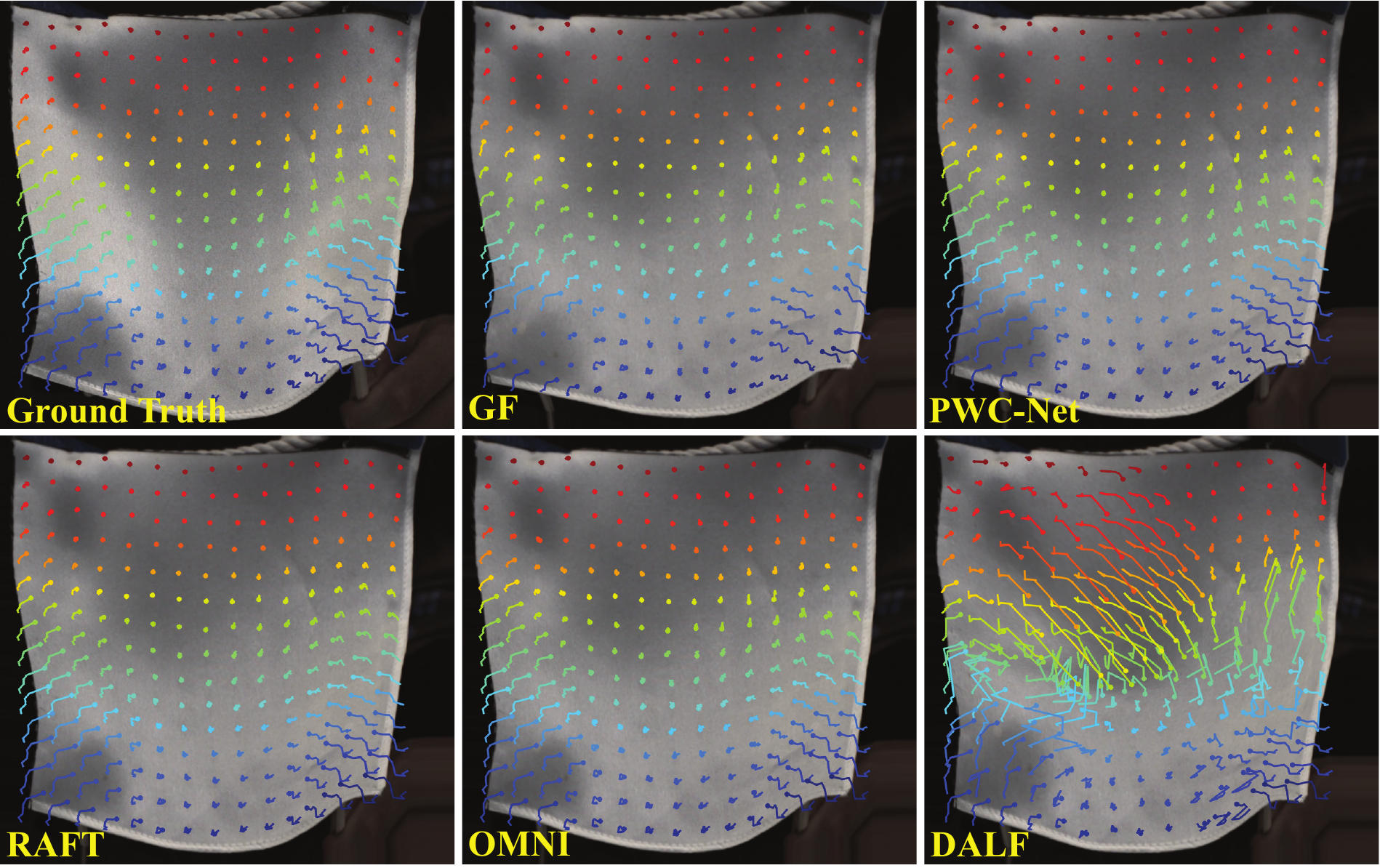}
\caption{Qualitative comparisons on warped images with accumulated flow trajectory (cloth scene $C4$). }
\label{fig:compare_opt_cloth}
\end{figure}

We visualize the optical flow estimation results by using the dense optical flow to warp the deformed frames. Results on several challenging paper and cloth scenes are shown in Fig.~\ref{fig:compare_opt_paper} and \ref{fig:compare_opt_cloth}, respectively. We also show the accumulated flow trajectory of 11 frames. We can see that most algorithms have large errors in textureless regions.

We further verified the accuracy of the warped images based on the grid corners. For each frame, we warp the image using the estimated flows on grid corners. Since the ground-truth warped frame image should be the next frame, we then compare the warped image to the image of the next frame and calculate the averaged root mean squared error (RMSE) for all frames in the motion sequence. Results on several cloth and paper scenes are reported in Table~\ref{tab:benchmark2_paper}. 

\begin{table}[t]
\centering
\footnotesize 
    \caption{Quantitative comparisons of warping errors on cloth and paper scenes.}    
    \label{tab:benchmark2_paper}
    \setlength\tabcolsep{5.5pt}
    \begin{tabular}{cccccccc} 
        \hline
        \multirow{2}{*}{Method} & \multicolumn{7}{c}{RMSE on warped image (\textbf{Cloth})} \\
        & $C2$ & $C3$ & $C4$ & $C5$ & $C9$ & $C10$ & Avg.\\ 
        \hline
        GF\cite{farneback2003two}
                              & 4.52 & 1.63 & 4.82 & 5.45 & 7.51 & 7.76 & 5.28 \\\cline{2-8}
        PWC\cite{sun2018pwc}
                              & 4.75 & 1.67 & 5.21 & 5.33 & 7.60 & 9.38 & 5.66 \\\cline{2-8}
        RAFT\cite{teed2020raft}
                              & 4.98 & 1.67 & 5.20 & 5.51 & 7.56 & 9.20 & 5.69 \\\cline{2-8}
        OMNI\cite{wang2023omnimotion}
                              & 4.89 & 1.65 & 5.52 & 5.60 & 7.71 & 8.93 & 5.72 \\\cline{2-8}
        DALF\cite{potje2023enhancing}
                              & 4.73 & 2.09 & 5.10 & 5.19 & 7.32 & 6.47 & 5.15 \\
        \hline
        \hline
        \multirow{2}{*}{Method} & \multicolumn{7}{c}{RMSE on warped image (\textbf{Paper})} \\
        & $P1$ & $P2$ & $P4$ & $P5$ & $P7$ & $P8$ & Avg.\\ 
        \hline
        GF\cite{farneback2003two}
                              & 8.96 & 11.2 & 7.91 & 10.5 & 11.0 & 11.7 & 10.2 \\\cline{2-8}
        PWC\cite{sun2018pwc}
                              & 13.8 & 6.89 & 4.34 & 4.85 & 4.59 & 6.38 & 6.89 \\\cline{2-8}
        RAFT\cite{teed2020raft}
                              & \color{black!20}{104*} & 7.40 & 4.59 & 5.61 & 5.36 & 6.12 & 5.87 \\\cline{2-8}
        OMNI\cite{wang2023omnimotion}
                              & 7.40 & 10.2 & 5.10 & 6.63 & 6.12 & 9.18 & 7.40 \\\cline{2-8}
        DALF\cite{potje2023enhancing}
                              & 9.44 & 14.8 & 4.59 & 5.61 & 4.85 & 7.40 & 7.91 \\
        \hline
    \end{tabular}
    
     \scriptsize{*This error is excluded for calculating the average.}
\end{table}

\noindent\textbf{Hand Reconstruction Methods.}
We test on four state-of-the-art template-based hand reconstruction methods: InterNet~\cite{Moon_2020_ECCV_InterHand2.6M}, IntagHand~\cite{Li2022intaghand}, InterWild~\cite{moon2023interwild}, and DIR\cite{ren2023decoupled}. All these methods take a single RGB image as input and predict the corresponding hand template using the MANO model~\cite{MANO:SIGGRAPHASIA:2017}. All methods are trained on InterHand2.6M~\cite{Moon_2020_ECCV_InterHand2.6M}. We use our front-view marker-free images as input to these methods. We project the 3D reconstruction back to the original 2D image plane and compute the per-pixel error on the wrapped image. Results on six challenging hand scenes are reported in Table~\ref{tab:hand_benchmark1}.

\begin{table}[t]
\centering
\footnotesize 
    \caption{Quantitative comparisons of hand reconstruction errors.}    
    \label{tab:hand_benchmark1}
    \setlength\tabcolsep{2.5pt}
    \begin{tabular}{cccccccc} 
        \hline
        \multirow{2}{*}{Method} & \multicolumn{7}{c}{RMSE on warped image (\textbf{Hand})} \\
        & $H1$ & $H2$ & $H3$ & $H4$ & $H5$ & $H6$ & Avg.\\ 
        \hline
        InterNet\cite{Moon_2020_ECCV_InterHand2.6M}
                              & 42.46 & 28.23 & 82.89 & 41.92 & 31.13 & 50.39 & 41.92 \\\cline{2-8}
        IntagHand\cite{Li2022intaghand}
                              & 34.88 & 27.02 & 64.09 & 28.30 & 30.91 & 54.64 & 34.89 \\\cline{2-8}
        InterWild\cite{moon2023interwild}
                              & 24.01 & 18.80 & 33.61 & 14.75 & 14.73 & 18.07 & 17.75 \\\cline{2-8}
        DIR\cite{ren2023decoupled}
                              & 31.57 & 27.73 & 50.42 & 23.32 & 28.44 & 44.88 & 32.16 \\\cline{2-8}
        \hline
        \hline
        \multirow{2}{*}{Method} & \multicolumn{7}{c}{Max error on warped image (\textbf{Hand})} \\
        & $H1$ & $H2$ & $H3$ & $H4$ & $H5$ & $H6$ & Avg.\\ 
        \hline
        InterNet\cite{Moon_2020_ECCV_InterHand2.6M}
                              & 111.32 & 69.42 & 166.00 & 123.41 & 95.51 & 124.27 & 97.01 \\\cline{2-8}
        IntagHand\cite{Li2022intaghand}
                              & 79.25 & 88.97 & 212.81 & 109.85 & 93.27 & 262.17 & 135.57 \\\cline{2-8}
        InterWild\cite{moon2023interwild}
                              & 67.10 & 61.14 & 116.05 & 40.49 & 47.98 & 63.62 & 62.80 \\\cline{2-8}
        DIR\cite{ren2023decoupled}
                              & 96.39 & 84.66 & 153.23 & 132.78 & 79.25 & 105.44 & 103.16 \\\cline{2-8}
        \hline
    \end{tabular}
\end{table}

\noindent\textbf{Deformable Object Tracking and Reconstruction.}

\begin{figure}[t]
\centering
\includegraphics[width=0.9\linewidth]{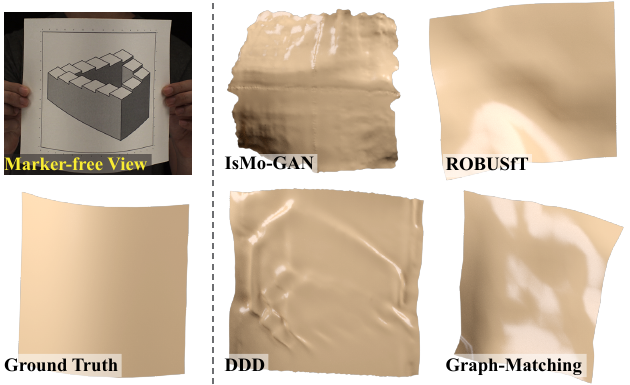}
\caption{Qualitative comparison on template estimation. The ground truth is reconstructed using our approach and provided in the dataset.}
\label{fig:mesh_compare}
\end{figure}

We test on four template-based deformable object tracking methods: DDD\cite{Yu2015DDD}, Graph-Matching\cite{Wang_2019_ICCV}, RoBuSfT\cite{shetabbushehri2023robusft}, and IsMo-GAN\cite{Shimada_2019}. The first three are optimization-based methods, and the last one is learning-based. These methods take in the image and template of a reference frame and output a deformed template for a target frame. We test these methods on marker-free views, evaluating the accuracy of the estimated template by computing vertex-to-vertex distance error against our ground truth 3D model. The errors are reported in Table \ref{tab:benchmark3}. Note that the results of IsMo-GAN cannot be evaluated in this way as their templates cannot be aligned with ours and have different scales. We visualize estimated deformed templates for one frame in $P2$ (see Fig.~\ref{fig:mesh_compare}). We can see that all these methods suffer from large errors on weakly textured paper scene.

\begin{table}[t]
\centering
\footnotesize 

    \caption{Quantitative comparisons on template reconstruction errors.} 
    \label{tab:benchmark3} 
    \setlength\tabcolsep{4pt}
    \begin{tabular}{cccccccc} 
        \hline
        \multirow{2}{*}{Method} & \multicolumn{7}{c}{Vertex-to-Vertex Distance Error} \\
        & $P1$ & $P2$ & $P4$ & $P5$ & $P7$ & Avg. &\\ 
        \hline
        {DDD\cite{Yu2015DDD}} 
                              & 25.22 & 15.14 & 28.50 & 24.28 & 23.75 & 23.44 \\\cline{2-8}
        {Graph-Match\cite{Wang_2019_ICCV}} 
                              & 379.77 & 115.32 & 95.35 & 18.48 & 24.16 & 126.62 \\\cline{2-8}
        {RoBuSfT\cite{shetabbushehri2023robusft}} 
                              & NAN & 60.31 & NAN & 2.64 & NAN & 31.48\\\cline{2-8}
        \hline
    \end{tabular}
\end{table}

\subsection{DOT Fine-tuned Network Results}
\label{sec:finetune}
\begin{figure*}[ht]
    \centering
    \includegraphics[width=0.9\linewidth]{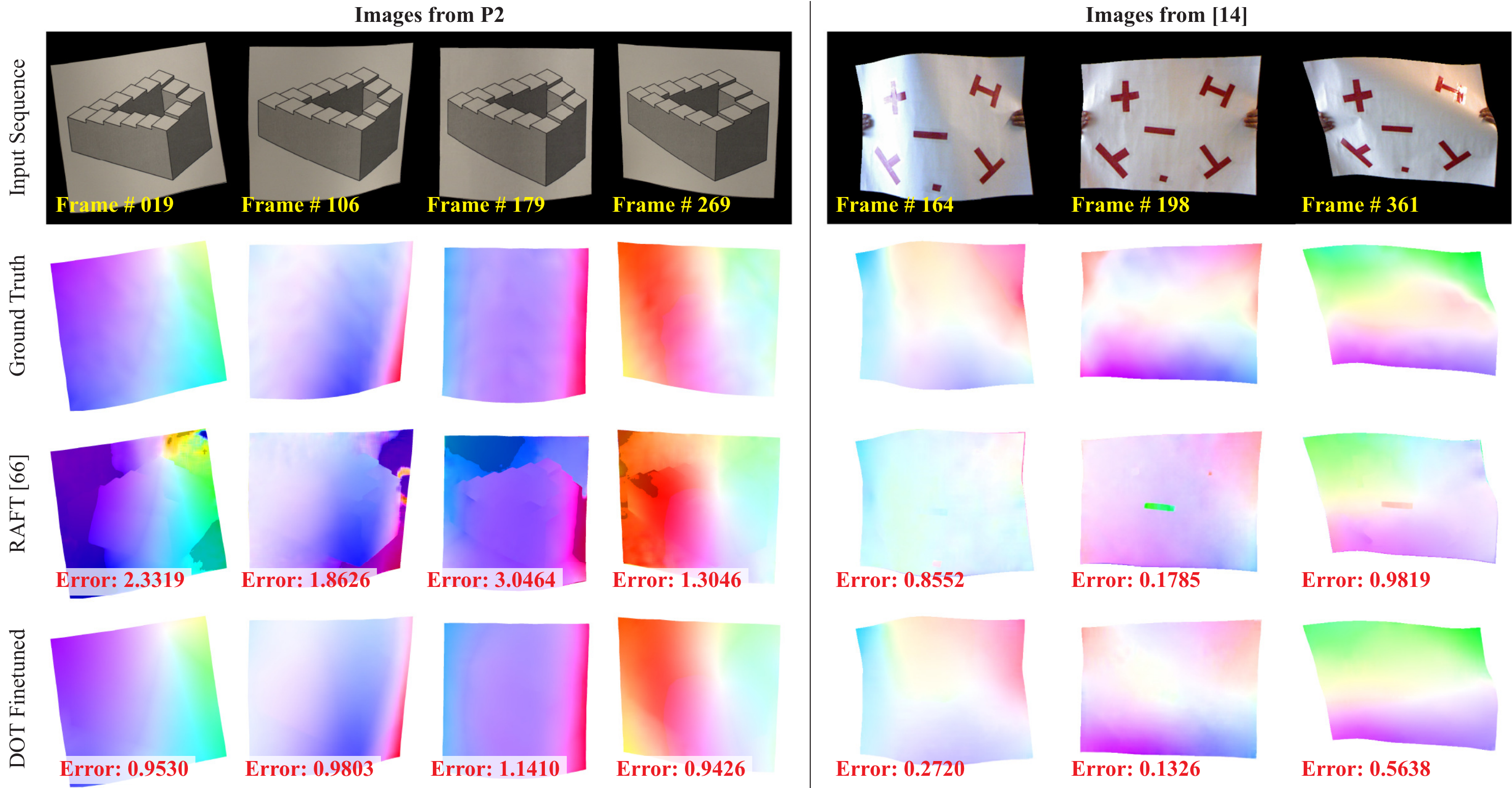}
    \caption{Qualitative comparisons on dense optical flow (pre-trained RAFT vs. fine-tuned RAFT). \textbf{(left)} $P2$ in DOT. \textbf{(right)} Data from DeSurT~\cite{Wang_2019_ICCV}.}
    \label{fig:finetune1}
\end{figure*}

Finally, we demonstrate the benefit of our DOT dataset on network training. Specifically, we test on optical flow and 3D template estimation. For optical flow, we use RAFT pre-trained on Flying Chairs~\cite{dosovitskiy2015flownet}. We then fine tune the pre-trained RAFT on our paper scenes. We apply the fine-tuned RAFT on $P2$ in DOT (not used in training), as well as data from the DeSurT dataset~\cite{Wang_2019_ICCV}, which provides a variety of challenging deformable surfaces with ground truth meshes. Dense optical flow results are shown in Fig.~\ref{fig:finetune1}. Here we compare the results of fine-tuned RAFT against the pre-trained RAFT. The ground truth flow maps are interpolated from the sparse flow at grid points. We also report the average flow errors. We can see that the accuracy of fine-tuned RAFT is significantly higher on all cases.

\begin{figure}[t]
\centering
\includegraphics[width=0.85\linewidth]{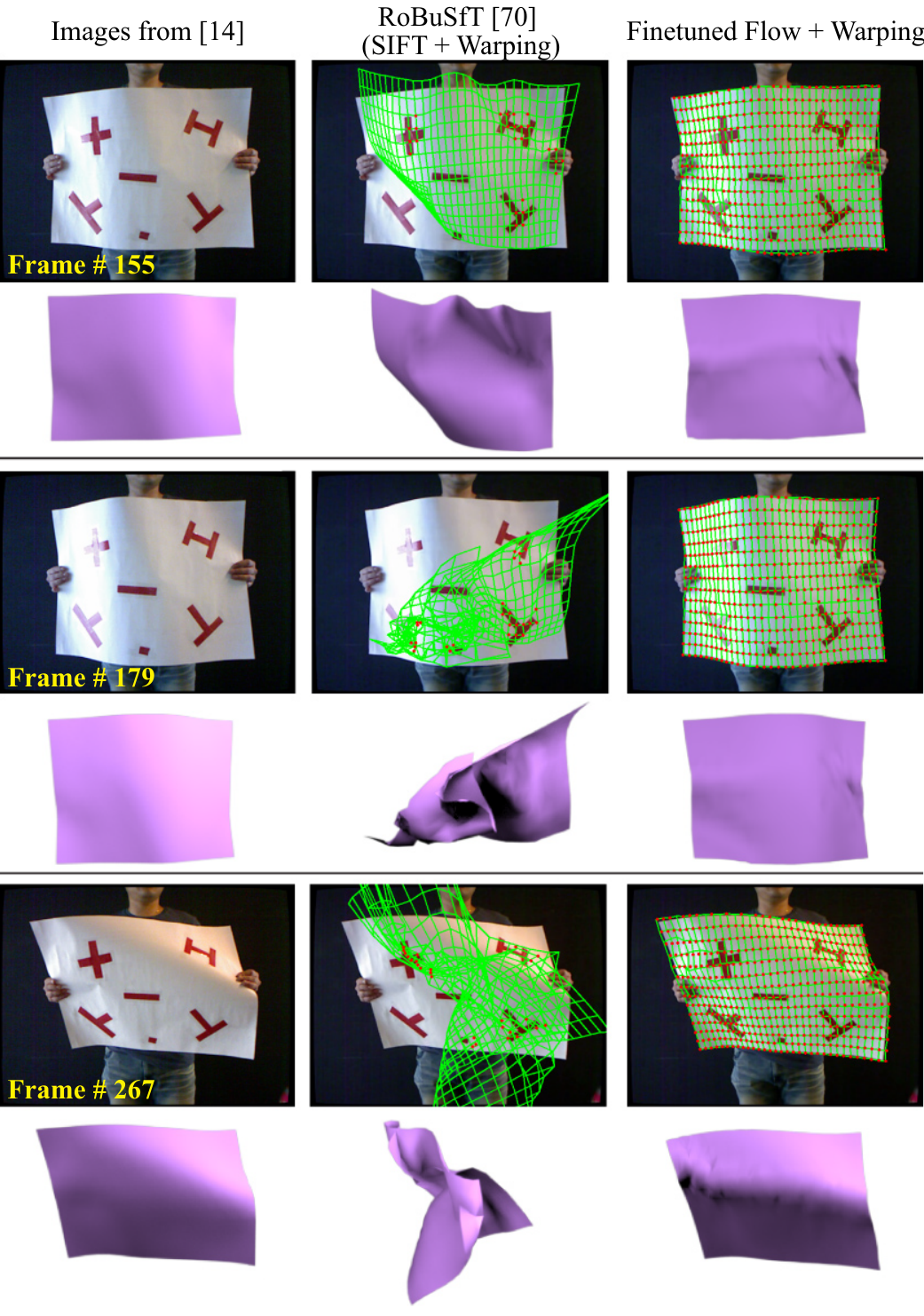}
\caption{Qualitative comparisons on template reconstruction with vs. without using our fine-tuned flow.}
\label{fig:warping_finetune}
\end{figure}

We then combine the fine-tuned flow correspondences with RoBuSfT~\cite{shetabbushehri2023robusft} for 3D template estimation. Since RoBuSfT uses SIFT for feature mapping by default, it fails at textureless regions. Whereas the optical flow fine-tuned on DOT is able to provide reliable correspondences regardless of textures, which allows accurate warping and template estimation. Comparison results on template reconstruction are shown in Fig.~\ref{fig:warping_finetune}.

\section{Conclusions}

In summary, we demonstrated a solution that uses invisible fluorescent markers for tracking and reconstructing deformable object with little texture. In contrast to existing methods, we are able to simultaneous capture videos of deformable object with and without markers. Videos with markers are used for accurate 3D reconstruction and feature tracking. Tracked correspondences can be transferred to the marker-free videos as ground truth labels. We collected a large deformable motion dataset, DOT, with 200 motion sequences and 1M video frames. DOT provides diverse forms of data, including multi-view videos with and without markers, 3D models and point clouds, and ground truth correspondences in both 2D and 3D. It can be used for benchmarking, or training networks for improved accuracy and robustness on textureless scenes.

\ifpeerreview \else
\section*{Acknowledgments}
The authors would like to thank...
\fi

\bibliographystyle{IEEEtran}
\bibliography{main}

\ifpeerreview \else






\fi

\end{document}


\maketitle

\appendices
In this supplementary document, we provide additional information about our UV florescent dye (Sec.~\ref{sec:dye_add}) and our acquisition system (Sec.~\ref{sec:sys_add}). We also include sample data in our DOT dataset (Sec.~\ref{sec:exp_add}). Please see the supplementary video for sample motion sequences in our DOT dataset. 









\section{Additional Information on Fluorescent Dye}
\label{sec:dye_add}
\subsection{Spectral Response}
\begin{figure}[h]
\centering
    \includegraphics[width=0.9\linewidth]{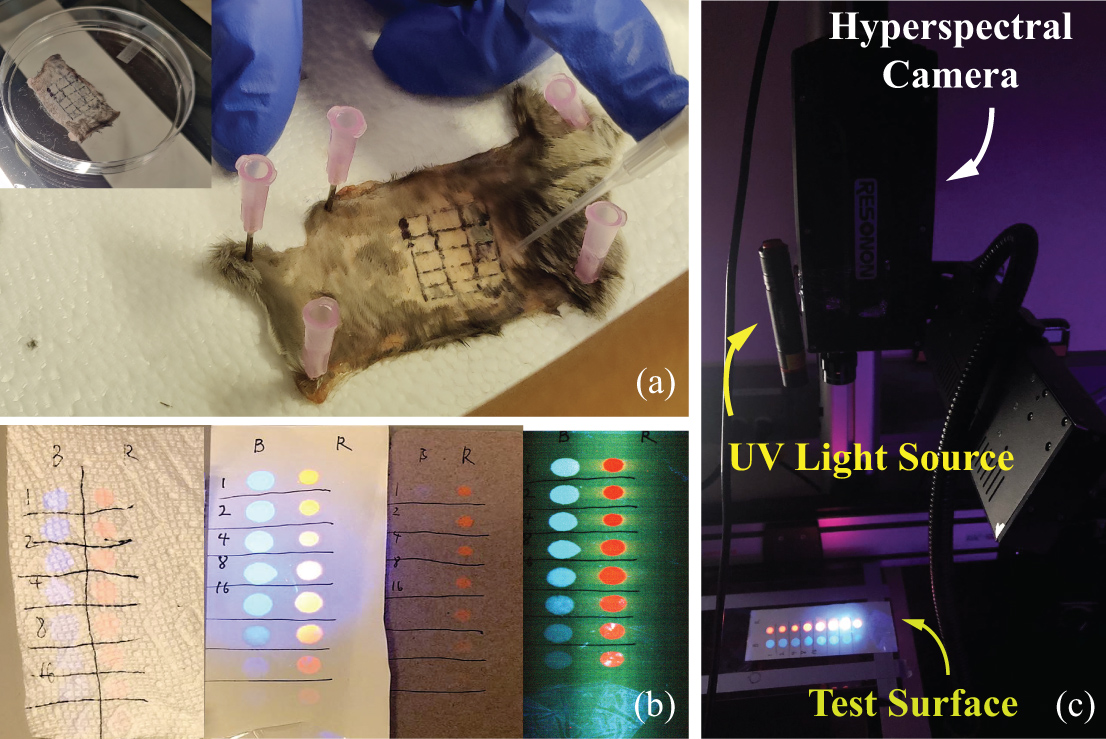}
    \caption{We measure the fluorescent spectral response on different types of surfaces. \textbf{(a)} Mouse skin sample. \textbf{(b)} Various paper and cloth samples. \textbf{(c)} Device for measuring the spectral response.}
    \label{fig:uv_test1}
\end{figure}

\begin{figure}[h]
\centering
    \includegraphics[width=1.0\linewidth]{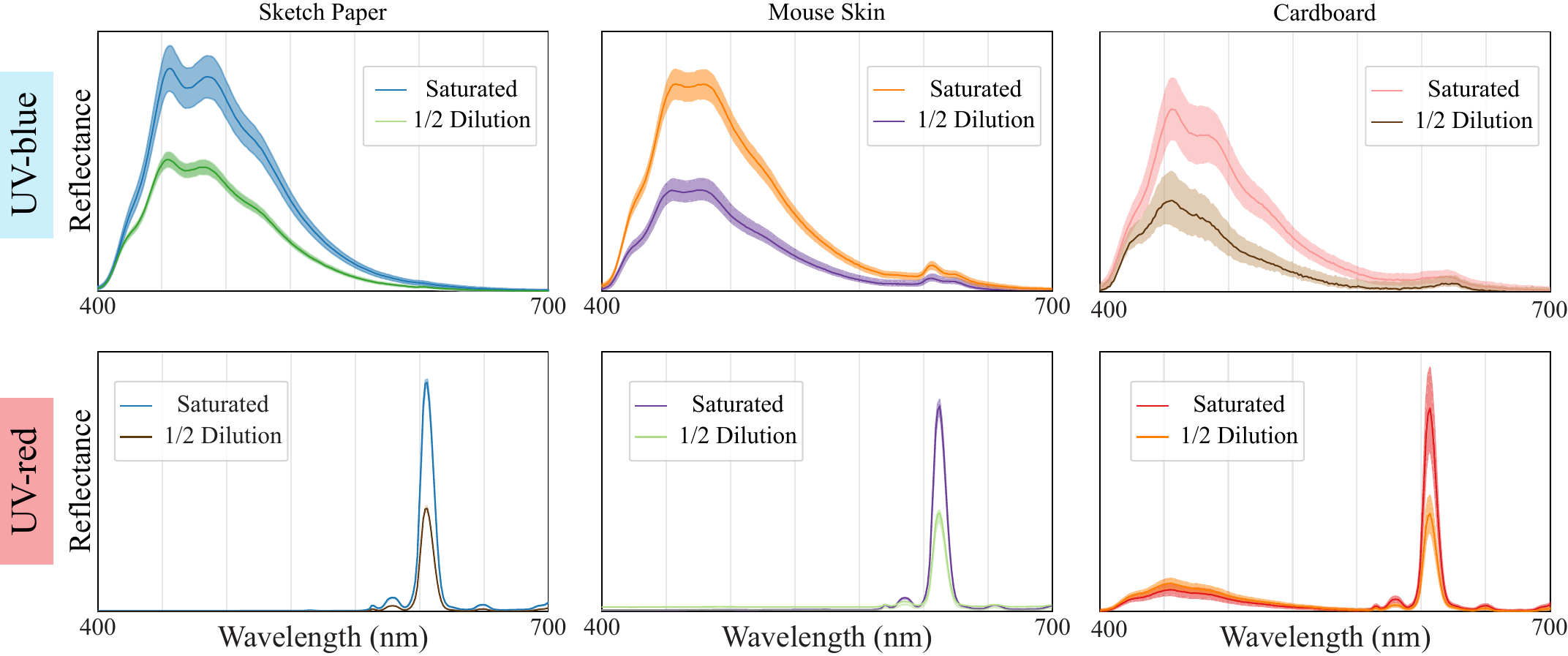}
    \caption{Spectral responses of two types of 
    markers on sketch paper (left), mouse skin (middle) and cardboard (right).}
    \label{fig:uv_test2}
\end{figure}

We measure the spectral response of our markers painted on different types of
surfaces, including the skin of a mouse (after hair removal), a piece of paper, and a cotton cloth. 
We include a mouse skin (kindly provided by a bio-medical lab)
because its composition is very close to the human skin.
We use a hyperspectral camera (Pika XC2) to
measure the spectral response of fluorescent emission under $365nm$ UV light (see Fig.~\ref{fig:uv_test1}).
The spectral response curves are shown in Fig.~\ref{fig:uv_test2}. 
The results indicate that each type of markers has similar spectral
response on various types of surfaces,
suggesting that our hue-based marker detection algorithm is applicable to 
a wide range of surfaces.

\subsection{Markers on Hand-object Interaction Scenes}

\begin{figure}[h]
\centering
\includegraphics[width=0.85\linewidth]{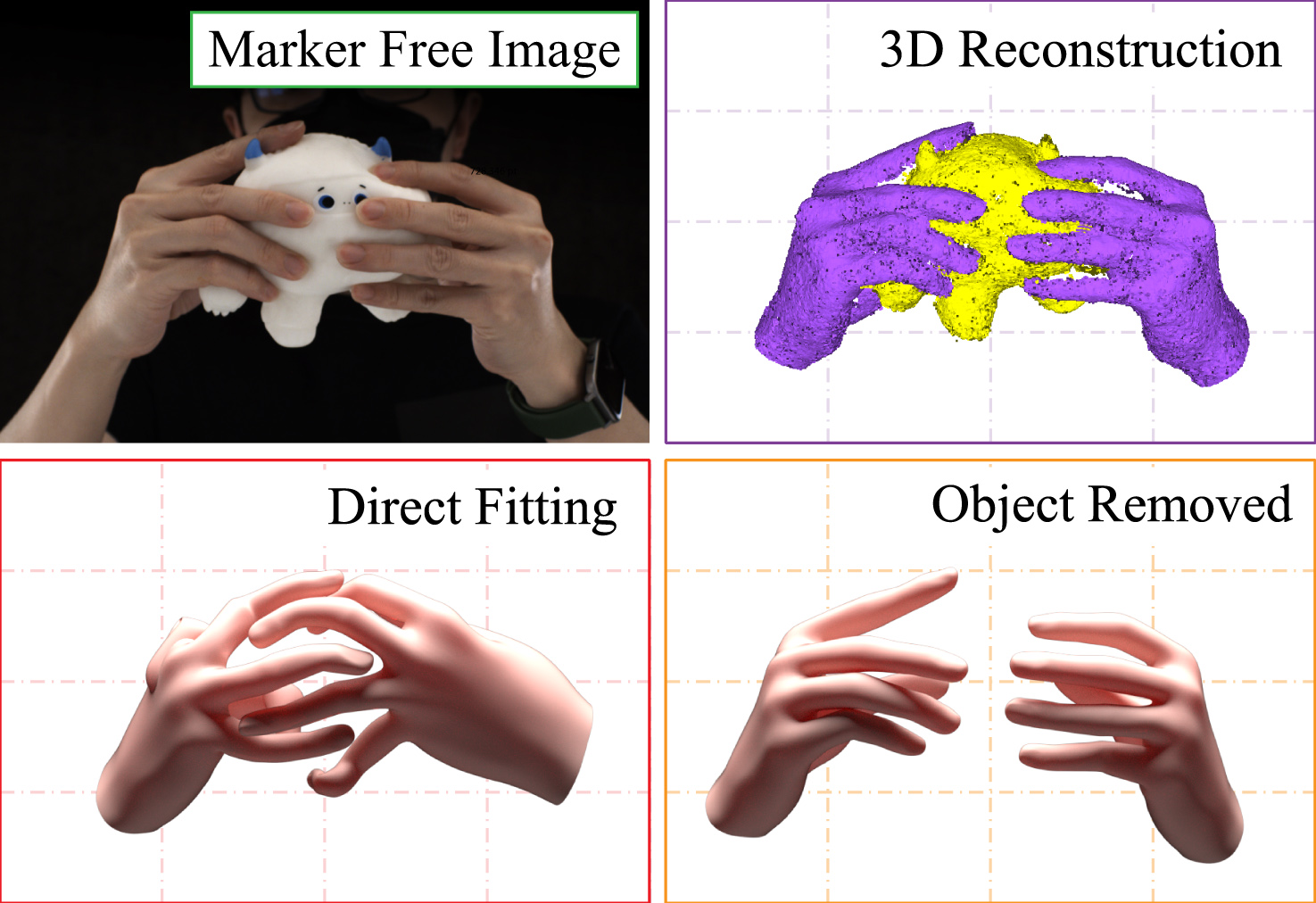}
\caption{Comparison of hand template fitting results with vs. without separating the object point cloud.}
\label{fig:hand_w_object}
\end{figure}

For most of the scenes that involve only one object, one type of fluorescent dye is sufficient for making the markers. For hand-object interaction scenes that involves two types of objects, we use both types of fluorescent dyes: we use UV-blue ink to draw dot patterns on the hand, and UV-red ink to fully cover the object. The two types of dyes help us separate the hand and object, which can significantly improve the template fitting. Fig.~\ref{fig:hand_w_object} shows a comparison of hand template fitting results with vs. without separating the object in hand. We can the direct fitting result (\ie, without separating the object) is prone to large errors, since the object points are also being considered in the fitting. However, by using different dye color to separate the hand and object, the fitting result is significantly improved. 

\subsection{Biological Safety}
\label{sec:safety}
Since we will use our system to capture human hands, we address the
safety concerns carefully.  First, the fluorescent dyes we use are non-toxic
and stable under non-direct sunlight. 
When applying markers on human skin, we only use the UV-blue dye dissolved in
70\% alcohol, because the alcohol solvent, commonly used for medical sterilization, 
is safe for skin contact.
The UV-red dye is dissolved in acetone, which is commonly used as nail polish
remover. But the acetone may irritate human skin.
Therefore, we do not apply it on human skin and only use it on
inanimate objects (such as paper and cloth). 

UV light is present in natural sunlight---it constitutes about 10\% of the total electromagnetic radiation from the sun, and about 95\% of UV
light is in the UVA spectrum (\ie, from $315~\mathrm{nm}$ to $400~\mathrm{nm}$) \cite{solar}. 
But long-term exposure or short-term over-exposure
to UV light can cause potential harm to human skin and eyes. To prevent from the harm, in our
experiments, we strictly follow the Threshold Limit Values and Biological
Exposure Indices (TLV/BEI) guidelines~\cite{TVI_BEI} to limit the UV
illumination. 

When we use our imaging system to acquire human hand gestures or faces, 
a typical acquisition session lasts no more than 10 minutes, during which
the UV lights are turned on for only 36 seconds. 
We measure using a spectrometer the UV irradiance when our imaging system is in use, and the average UV irradiance is $24~\mathrm{\mu W/cm^2}$. As a comparison,
the UV irradiance on a bright sunny day under direct sun light is about $250~\mathrm{\mu
W/cm^2}$ \cite{solar}, that is, around ten times of our UV light irradiance. 
Therefore, we believe our system will not cause any UV over-exposure risk.  
To be extra cautious, we always apply sunscreen cream on human hands and faces
before painting the markers and exposing them to our UV
lights.  Also, our system operator always wears
a UV protective glass with Optical Density (OD) $>6$ during the acquisition
process. In sum, our markers and UV lights cause no significant
safety hazard, and we have taken additional safety measures for protection. 

\section{Additional Information on Camera System} 
\label{sec:sys_add}
\subsection{Camera and Light Source Specs}

All cameras capture videos with a resolution of $2448\times2048$ at
$60~\mathrm{fps}$. They have a 30-degree field of view and an aperture of $f/5.6$ 
in order to capture all-focus images within the range of the rig ($\approx 40~\mathrm{cm}$).
In addition, we mount a $400~\mathrm{nm}$ long-pass filter in front of each cameras. 
The filter blocks the visible blue light component caused by the UV lights from
contaminating the captured images.



Each UV LED unit has a power of $2.58~\mathrm{W}$, emitting light with a peak
wavelength at $365~\mathrm{nm}$ and a beam angle of $155^\mathrm{\circ}$. 
The UV LEDs response rapidly to the triggering voltage:
their rise and fall times are within $1~\mathrm{ns}$.
In addition, we use 7 
light panels on the top and bottom sides of the frame. These light panels
are always on during a capture session in order to supply more light such that
we can use a faster shutter speed and reduce captured motion blur and surface specularity. 

\subsection{Calibration \& Data Management} 
\label{sec:calib}
All cameras are geometrically calibrated 
using ChArUco patterns attached to each face of a rhombicuboctahedron. 
We also estimate the intrinsic and extrinsic parameters of all cameras, from which
we compute the geometric transformation between each pair of cameras.
This allows us to re-project images captured by one camera to another camera 
(with the re-projection error less than 0.6 pixels).
We chromatically calibrate all cameras through adjusting their white balance.

With 42 cameras capturing 2K images at $60~\mathrm{fps}$, we need an
infrastructure to support high-speed data transfer and storage. 
To this end, we set up three high-performance servers, each with $16~\mathrm{TB}$ SSD hard drive and PCI-E to USB capture cards. 
On each server, we build an M2 SSD RAID system with read and sequential write
speeds up to $28~\mathrm{Gbps}$, and connect it to 14 cameras.
In this way, we can directly save uncompressed raw images on the servers. 
The RAID systems can host up to 1 hour of acquired images. 

\begin{figure*}[h]
\centering
\includegraphics[width=1\linewidth]{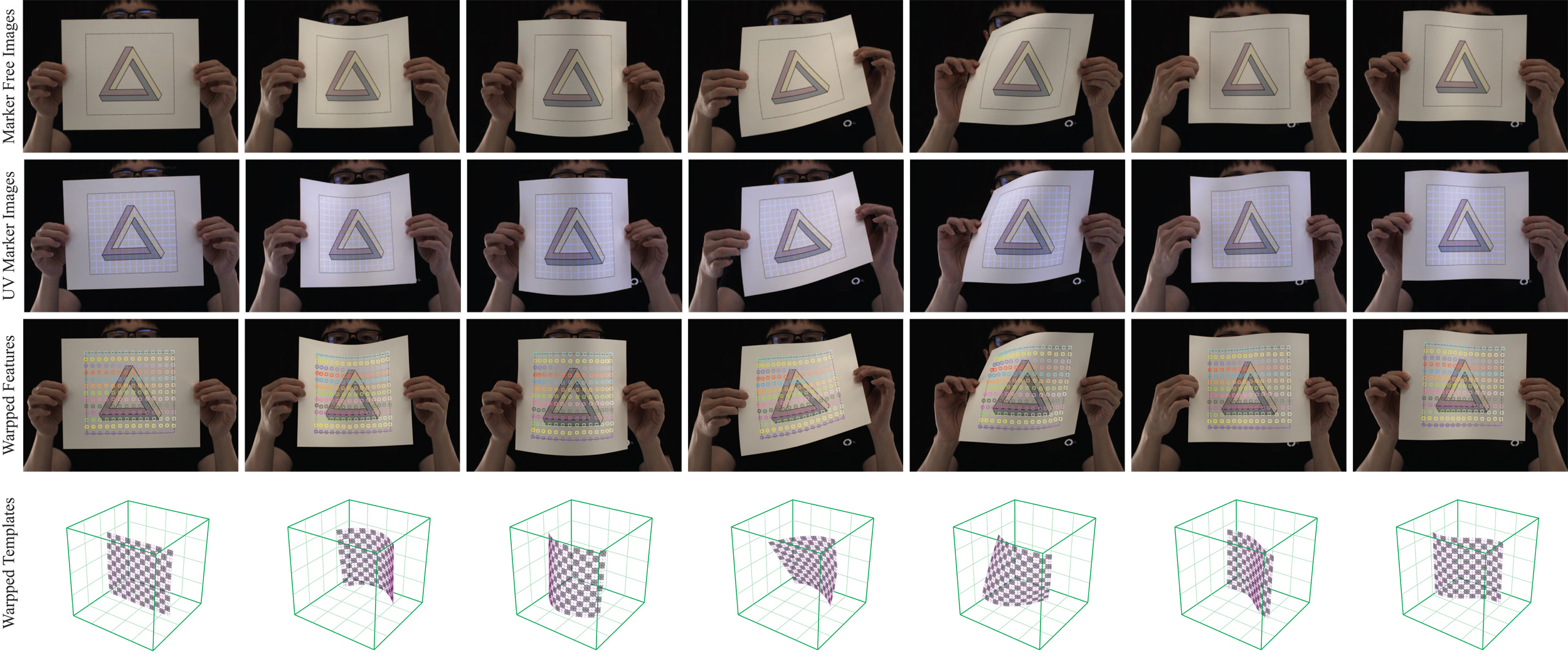}
\caption{\textbf{Deformed paper}. From top to bottom, we show marker-free images, UV images with markers, tracked color-coded correspondences, and recovered 3D surface.}
\label{fig:deform}
\end{figure*}

\begin{figure*}[h]
\centering
\includegraphics[width=1\linewidth]{Fig/hand_res_2.pdf}
\caption{\textbf{Hand interactions}. From top to bottom, we show marker-free images, UV images with markers, recovered 3D hand and object point clouds, and recovered 3D hand meshes warped on the marker-free views.}
\label{fig:hands}
\end{figure*}


\section{Sample Data from DOT}
\label{sec:exp_add}
Here we show some sample data from our DOT dataset. Specifically, Fig.~\ref{fig:deform} shows images from one sequence of a paper scene ($P9$). Here we show marker-free view, UV view with markers, marker-free view with warped feature correspondences, and 3D template. Fig.~\ref{fig:hands} shows example hand interaction scenes, which include two-hand interaction and hand-object interaction. Please see the supplementary video for more examples from all four categories and with various motions.

\bibliographystyle{IEEEtran}
\bibliography{main}